\documentclass{article}
\usepackage[utf8]{inputenc}

\usepackage{tikz,pgfplots}
\pgfplotsset{compat=1.18}
\usetikzlibrary{positioning, arrows, shapes.geometric, calc, decorations, decorations.pathreplacing, decorations.text, backgrounds,fit,spy,patterns, shapes,}

\usepackage{xr}
\usepackage{geometry}
\newenvironment{linenomath*}{}{}
\title{Deep unfolding of MCMC kernels: scalable, modular \& explainable GANs for high-dimensional posterior sampling}

\author{
	Jonathan Spence\footnote{Maxwell Institute for Mathematical Sciences, Bayes Centre, 47 Potterrow, Edinburgh, Scotland, UK} \thanks{School of Mathematical and Computer Sciences, Heriot-Watt University,  Edinburgh, Scotland, UK}
	\and
	Tob\'ias I. Liaudat\thanks{IRFU, CEA, Universit\'e Paris-Saclay, F-91191 Gif-sur-Yvette, France}
	\and Konstantinos Zygalakis\footnotemark[2] \thanks{School of Mathematics, University of Edinburgh,  Edinburgh, Scotland, UK}
	\and Marcelo Pereyra\footnote{Corresponding author. E-Mail \href{M.Pereyra@hw.ac.uk}{M.Pereyra@hw.ac.uk}} \footnotemark[2] \footnotemark[3]
}

\usepackage{makecell}
\usepackage{multirow}
\usepackage{colortbl}
\usepackage{amsmath, amsfonts, amssymb}
\usepackage[nocompress]{cite}
\usepackage{hyperref}
\usepackage{cleveref}
\usepackage{subcaption}
\usepackage{algorithmic}
\crefname{algorithm}{Algorithm}{Algorithms}
\usepackage{bbm}
\date{}
\def\mywidth{1.75}
\def\mywidthcm{\mywidth cm}
\newif\ifTodo
\Todotrue 

\ifTodo
\newcommand{\todo}[1]{{\bf\color{red} {#1}}}
\else 
\newcommand{\todo}[1]{}
\fi

\def\rset{\mathbb{R}}
\def\dx{d_x}
\def\xspace{\rset^{\dx}}
\def\dy{{d_y}}
\def\yspace{\rset^{\dy}}
\def\noise{\epsilon}
\def\op{A}
\def\id{I}
\DeclareMathOperator{\prox}{prox}

\usepackage{algorithm}
\usepackage{ifthen}

\definecolor{myblue}{RGB}{8, 54, 193}

\usepackage{amsmath,amsfonts,bm}

\def\1{\bm{1}}

\def\rvk{{\mathbf{k}}}
\def\rvl{{\mathbf{l}}}
\def\rvm{{\mathbf{m}}}

\def\rvw{{\mathbf{w}}}
\def\rvx{{\mathbf{x}}}
\def\rvy{{\mathbf{y}}}
\def\rvz{{\mathbf{z}}}

\DeclareMathAlphabet{\mathsfit}{\encodingdefault}{\sfdefault}{m}{sl}
\SetMathAlphabet{\mathsfit}{bold}{\encodingdefault}{\sfdefault}{bx}{n}

\newcommand{\E}{\mathbb{E}}

\newcommand{\keywords}[1]{\par\mbox{}\par\noindent\textbf{Keywords:} #1}

\DeclareMathOperator*{\argmax}{arg\,max}
\DeclareMathOperator*{\argmin}{arg\,min}

\def\visdir{visualizations}

\begin{document}
\maketitle
\begin{abstract}
Markov chain Monte Carlo (MCMC) methods are fundamental to Bayesian computation, but can be computationally intensive, especially in high-dimensional settings. Push-forward generative models, such as generative adversarial networks (GANs), variational auto-encoders and normalising flows offer a computationally efficient alternative for posterior sampling. However, push-forward models are opaque as they lack the modularity of Bayes Theorem, leading to poor generalisation with respect to changes in the likelihood function. In this work, we introduce a novel approach to GAN architecture design by applying deep unfolding to Langevin MCMC algorithms. This paradigm maps fixed-step iterative algorithms onto modular neural networks, yielding architectures that are both flexible and amenable to interpretation. Crucially, our design allows key model parameters to be specified at inference time, offering robustness to changes in the likelihood parameters. We train these unfolded samplers end-to-end using a supervised regularized Wasserstein GAN framework for posterior sampling. Through extensive Bayesian imaging experiments, we demonstrate that our proposed approach achieves high sampling accuracy and excellent computational efficiency, while retaining the physics consistency, adaptability and interpretability of classical MCMC strategies.
\keywords{Bayesian statistics, inverse problems, deep unfolding, generative adversarial networks, Markov Chain Monte Carlo, computational imaging,}
\end{abstract}

\section{Introduction}
\label{sec:intro}
We consider the inverse problem of reconstructing an $\xspace$-valued signal $x^\star$ given an $\rset^\dy$-valued measurement $\rvy$ with a likelihood function
\begin{linenomath*}\begin{equation}
\label{eqn:inverse_problem}
\begin{aligned}
	\rvy &\sim p(y|x^\star) \propto \exp(-f_y(x^\star)),\end{aligned}
\end{equation}\end{linenomath*}
where the negative log-likelihood $f_y:\rset^{\dx}\to\rset$ is convex for all $y$. As a canonical example, we consider the linear Gaussian problem
\begin{linenomath*}\begin{equation}
\label{eqn:linear_inverse_problem}
\begin{aligned}
    y = \op x^\star + \noise,\qquad \noise \sim \mathcal N(0, \sigma_y^2\id_\dy),
\end{aligned}
\end{equation}
\end{linenomath*}
with likelihood $p(y|x) = \mathcal N(y; \op x, \sigma_y^2\id_\dy)$,  
where $\op:\xspace\to\yspace$ is a linear degradation operator and $\sigma_y>0$ controls the scale of measurement noise. Such problems arise in multiple application areas including compressive sensing, tomography \cite{Kaipio2010} and in computational imaging, with applications to scientific, medical and natural images \cite{Ongie2020}. The methods studied in this paper are intended for the general class of problems taking the form \eqref{eqn:inverse_problem}, including but not limited to those mentioned above. 

The inverse problem \eqref{eqn:inverse_problem} is typically ill-conditioned. For example, in the  canonical Gaussian example, $\op$ is often low-rank or has a high conditioning number. To obtain accurate reconstructions, we therefore require additional regularity constraints on the space of solutions. Within a statistical framework, we assume $x^\star$ is a realisation of the random variable $\rvx\sim p(x)$ for an (unknown) distribution $p(x)$ defined on $\xspace$. By Bayes Theorem \cite{robert2007bayesian}, the distribution of signals $\rvx$, given a measurement $\rvy$ is expressed through the posterior distribution $$p(x|y) = \frac{p(y|x)p(x)}{\int_{\rset^{dx}} p(y|x)p(x)\text{d}x}.$$

In practice, the exact distribution of $\rvx$ is unknown, precluding direct inference using $p(x|y)$. Instead, we approximate $p(x)$ by a parametric prior distribution $p_\theta(x)=\exp(-g_\theta(x))$, where $g_\theta:\xspace\to \rset$ depends on the parameter(s) $\theta\in\rset^{d_\theta}$ which are tuned depending on the problem. For ill-posed inverse problems, a choice of prior which is accurately able to describe the underlying data distribution of $\rvx$ is crucial. 
Traditionally, the $p_\theta$ would be designed to promote solutions with desired or expected regularity properties such as smoothness. Modern approaches for modelling prior knowledge are predominantly data-driven and use a deep neural-network with high-dimensional weights $\theta$ learned to fit an empirical data distribution.

Given a chosen parametrization of the prior, we consider the associated posterior distribution
\begin{linenomath*}\begin{equation}
	\label{eqn:bayesian}
	\pi_\theta(x|y)= \frac{p(y|x)p_\theta(x)}{\int_{\rset^{\dx}} p(y|x)p_\theta(x)}.
\end{equation}\end{linenomath*}
The proportionality constant in \eqref{eqn:bayesian} and properties of the posterior distribution (mean, variance) are typically unknown, necessitating numerical integration. Techniques from optimization can be used to maximise the posterior density, providing a point estimate $x_\text{MAP} = \argmax_x \pi_\theta(x|y)$. To obtain higher order statistical information, Markov Chain Monte Carlo (MCMC) techniques are often utilised to sample a Markov Chain ergodic with respect to $\pi_\theta(x|y)$. As $\ell\to \infty$, the $\ell$-step transition kernel of the Markov Chain converges to the posterior distribution. Approximate samples from \eqref{eqn:bayesian} can therefore be obtained from long-time integration of the Markov chain, after discarding a suitable burn-in period. These samples can then be used, for example, to inform statistical decision-making about the distribution of $\rvx$ given the observation $\rvy=y$.

Typically, the  weights $\theta$ are learned during an offline phase, following which a chosen numerical scheme is employed to sample $\pi_\theta(x|y)$. Thus $p_\theta$ is independent of the likelihood model $p(y|x)$. In the machine learning community, such methods are referred to as zero-shot: the task of reconstructing $x$ given the observation $y$ is performed without specialising the model for solving the inverse problem \eqref{eqn:inverse_problem} during an offline training phase. Alternatively, one can consider training $p_\theta$ specifically with the likelihood $p(y|x)$ and the chosen numerical scheme to optimise the exploration of the posterior \eqref{eqn:bayesian}.

MCMC with a data-driven prior provides a modular, transparent approach to Bayesian inverse problem reconstruction \cite{Holden2022,laumont2022bayesian,Mukherjee2023}.  Such  methods have been extensively studied in the imaging literature in recent years, particularly in combination with score-based generative diffusion models \cite{zhu2023denoising,song2021denoising,ho2020denoising}. However, such approaches often rely on a large number of numerical integrations approximating an underlying continuous-time process (for example Langevin or reversed-SDE), resulting in accurate but slow algorithms. Black-box methods for sampling $\pi_\theta(x|y)$ include conditional GANs \cite{adler2018,bendel2023regularized}, which train a deep neural-network adversarially end-to-end to transport $y$ into samples from $\pi_\theta(x|y)\approx p(x|y)$. Black-box models can be both accurate and fast, but come at the expense of interpretability lacking a natural embedding of the likelihood model $p(y|x)$ in the reconstruction process, making it difficult to account for small perturbations in the likelihood model at inference time.

In this work, we propose a method of combining data-driven MCMC and conditional GANs.  Specifically, we propose a novel framework for unfolding $L$ iterations of a truncated MCMC chain designed to target the true posterior $p(x|y)$. The proposed approach builds upon the unfolding of deterministic optimization schemes, a technique which jointly trains $\theta$ and hyper-parameters from $L$-iterations of a truncated numerical optimization scheme as a $L$-layer recurrent neural network, is considered in \cite{chen2022l2o,gregor2010learning} (see \Cref{fig:intro/unfolded_gd_diagram} and \Cref{sec:background}).   The proposed architecture is illustrated in \Cref{fig:intro/unfolded_mcmc_diagram}. The unfolded MCMC model provides high-order statistical information in contrast to optimization methods. This information is crucial, for example, in providing uncertainty estimation on model outputs. Leveraging the modularity and interpretability of Markov transition kernels, the proposed deep unfolded generative networks strike a careful balance between performance and explainability. Moreover, the posterior \eqref{eqn:bayesian} contains a natural embedding of the likelihood. Thus, an unfolded MCMC kernel targeting \eqref{eqn:bayesian} provides a natural embedding of the likelihood function, which is often lacking in black-box neural-network architectures trained for end-to-end reconstruction.  A recent and related work which considers the unfolding of a stochastic denoising diffusion sampler is \cite{mbakam2025learning}.

The paper is structured as follows: \Cref{sec:background} discusses background on Bayesian and data-driven methods for ill-posed inverse problems with a discussion of related literature. The proposed unrolled MCMC architecture and a training procedure tailored to such neural networks is discussed in \Cref{sec:methodology}. In \Cref{sec:numerics}, we benchmark unfolded MCMC architectures on image deblurring and radio interferometry problems. These problems illustrate two distinct forms of challenging inversion: the deblurring operator is severely ill-conditioned, whereas radio interferometry is ill-posed.

The key contributions of this work are summarised as follows.
\begin{itemize}
	\item A formal framework is proposed for general unfolded MCMC algorithms for posterior sampling. By considering a broad class of algorithms, we intend the methodology to be beneficial to a wide range of inverse problems.
	\item We propose a novel training scheme for GANs arising from unfolded MCMC kernels in \eqref{eqn:unfolded_mcmc}. To promote accurate sampling of posterior distributions, our approach is inspired by recent developments in conditional adversarial networks \cite{bendel2023regularized,adler2018}, distilled diffusion \cite{song23consistency} and unrolled optimization \cite{Monga2021}.
	\item A systematic empirical study of unrolled MCMC architectures for two problems: a comprehensive ablation study on the MNIST dataset to afford accurate comparison of posterior distributions; and a high-dimensional problem in radio interferometry, with gridded visibilities simulated visibility patterns from the MeerKAT radio telescope.
\end{itemize}

\section{Background, Notation and Related Works}
\label{sec:background}
This section introduces key notation and summarises key background information on techniques for computational Bayesian inverse problems. In what follows, we assume the existence of a set of empirical data $\mathcal X = \{x^{(n)}\}_{n=1}^{N_\text{data}}$ of samples from the intractable prior $p(x)$. Crucially, this allows one to consider data-driven parametrisations $p_\theta(x)$ leveraging techniques from machine learning.

\subsection{Setup and notation}
We use $p_{\rvx, \rvy}$, $p_\rvx$ and $p_\rvy$ to denote the joint law of $(\rvx,\ \rvy)$, the law of $\rvx$ and the law of $\rvy$, respectively. Given measurable spaces $(\Omega_1, \Sigma_1)$ and $(\Omega_2, \Sigma_2)$, a measurable function $f:\Omega_1\to\Omega_2$ and a measure $\mu:\Sigma_1\to \rset_+$, we denote the push-forward measure $f_\#\mu:\Omega_2\to\rset_+$.  For a random variable $\rvw\sim p_\rvw$ taking values in a Banach space $E$ and measurable function $f$, we denote $\E_{\rvw\sim p_\rvw}[f(\rvw)] = \int_E f(\rvw)\text{d}p_\rvw$. When the random variable to be integrated is obvious from context, we simplify the notation to  $\E[f(\rvw)] = \int_E f(\rvw)\text{d}p_\rvw$. For a convex function $g:\xspace\to\rset$, we define the proximal operator of $g$ as the function ${\prox_{g}(x) = \argmin_z g(z) + \|x-z\|^2/2}$. Given a (deep) artificial neural network $\text{NN}_\theta:y_{\text{in}}\mapsto x_{\text{out}}$, with weights $\theta$, by a neural function evaluation (NFE) we refer to a single call $\text{NN}_\theta(x_\text{in})$. For a parametrised posterior distribution, we denote a random variable with distribution $\pi_\theta(x|y)$ by $\rvx_\theta^\rvy$. The joint law of $(\rvx_\theta^\rvy, \rvy)$ is then denoted $p_{\rvx_\theta^\rvy, \rvy}$.

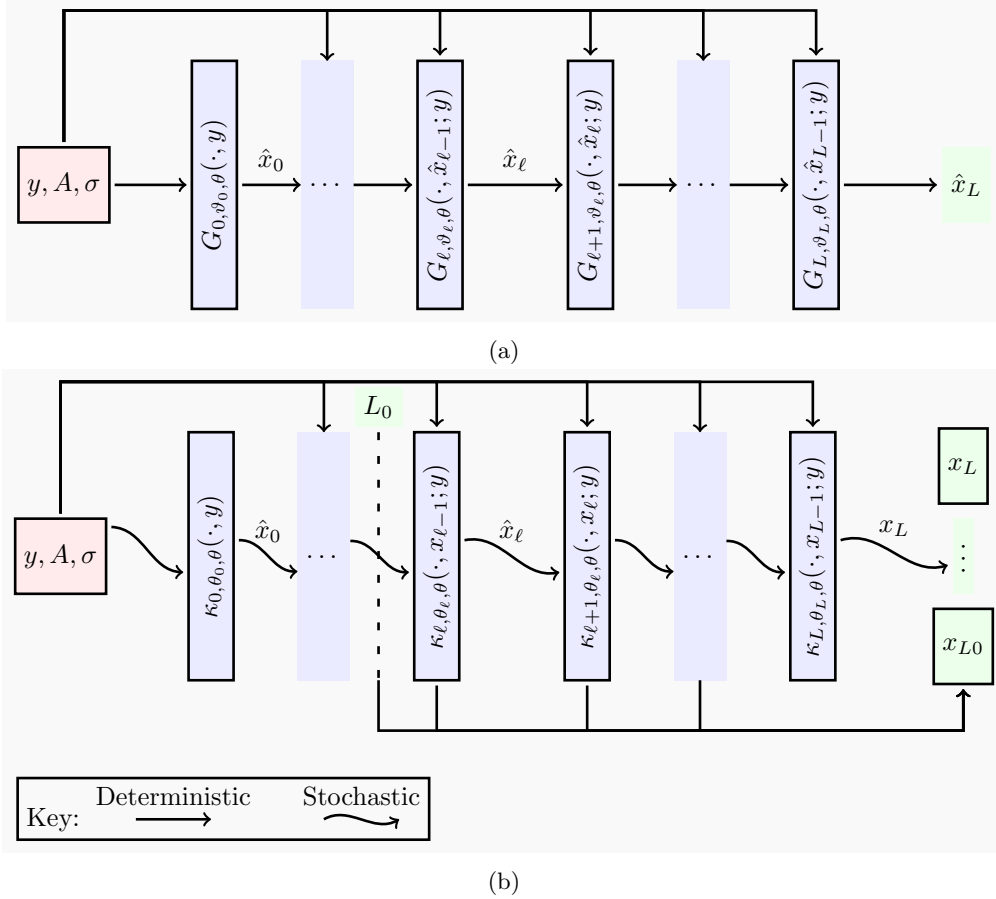
\begin{figure}
	\centering
	
	\begin{subfigure}{\linewidth}
	\centering
	\rotatebox{0}{
		\begin{tikzpicture}[node distance=2cm, minimum height=3.3cm, every edge/.style={inner sep=2pt, line width=6pt}, every node/.style={outer sep=2pt, fill=blue!8}, every path/.style={line width=1pt},show background rectangle, background rectangle/.style = {fill = gray!5,line width=1pt}]
		\node [draw, minimum height=1cm,fill=red!8] (i)  {\rotatebox{0}{$y,A,\sigma$}};
		\node [draw] (k0) [right of=i] {\rotatebox{90}{$ G_{0, \vartheta_0, \theta}(\cdot, y)$}};
		\node[node distance=1.5cm,outer sep=0pt] (kdot) [right of=k0] {\dots};
		\node[draw, node distance=1.5cm] (kl) [right of=kdot] {\rotatebox{90}{$G_{\ell, \vartheta_{\ell}, \theta}(\cdot, \hat x_{\ell-1};y)$}};
		\node[draw] (klp1) [right of=kl] {\rotatebox{90}{$G_{\ell+1, \vartheta_{\ell}, \theta}(\cdot, \hat x_{\ell};y)$}};
		\node[node distance=1.5cm,outer sep=0pt] (kdot2) [right of=klp1] {\dots};
		\node[draw, node distance=1.5cm] (kL) [right of=kdot2] {\rotatebox{90}{$G_{L, \vartheta_{L}, \theta}(\cdot, \hat x_{L-1};y)$}};
		\node[minimum height=1cm,fill=green!8] (out1) [right of=kL] {\rotatebox{0}{$\hat x_L$}};
		
		\draw [->]  (i) -- (k0);
		\draw [->]  (k0) -- node [above, minimum height=0.5cm,fill=none] {\rotatebox{0}{$\hat x_0$}} (kdot);
		\draw [->]  (kdot) --   (kl);
		\draw [->] (kl) -- node [above, minimum height=0.5cm,fill=none] {\rotatebox{0}{$\hat x_\ell$}} (klp1);
		\draw [->] (klp1) --  (kdot2);
		\draw [->] (kdot2) -- (kL);
		\draw [->] (kL) -- (out1);
		
\coordinate (above_start) at ($(k0.north) + (0, 0.6)$);
		\coordinate (below_start) at ($(kL.south) + (0, -0.6)$);
		\coordinate (disc) at ($(kL.south) + (3, -0.6)$);
		\draw [->] (i) |- (above_start) -| (kl);
		\draw [->] (i) |- (above_start) -| (kdot);
		\draw [->] (i) |- (above_start) -| (klp1);
		\draw [->] (i) |- (above_start) -| (kdot2);
		\draw [->] (i) |- (above_start) -| (kL);

	\end{tikzpicture}
	}
	\caption{} 
	\label{fig:intro/unfolded_gd_diagram}	
	\end{subfigure}
	\begin{subfigure}{\linewidth}
	\centering
	\rotatebox{0}{
	\begin{tikzpicture}[node distance=2cm, minimum height=3.3cm, every edge/.style={inner sep=2pt, line width=6pt}, every node/.style={outer sep=2pt, fill=blue!8}, every path/.style={line width=1pt},show background rectangle, background rectangle/.style = {fill = gray!5,line width=1pt}]
	\node [draw, minimum height=1cm,fill=red!8] (i)  {\rotatebox{0}{$y, A, \sigma$}};
	\node [draw] (k0) [right of=i] {\rotatebox{90}{$ \kappa_{0, \theta_0, \theta}(\cdot, y)$}};
	\node[node distance=1.5cm,outer sep=0pt] (kdot) [right of=k0] {\dots};
	\node[draw, node distance=1.5cm] (kl) [right of=kdot] {\rotatebox{90}{$\kappa_{\ell, \theta_{\ell}, \theta}(\cdot, x_{\ell-1};y)$}};
	\node[draw] (klp1) [right of=kl] {\rotatebox{90}{$\kappa_{\ell+1, \theta_{\ell}, \theta}(\cdot, x_{\ell};y)$}};
	\node[node distance=1.5cm,outer sep=0pt] (kdot2) [right of=klp1] {\dots};
	\node[draw, node distance=1.5cm] (kL) [right of=kdot2] {\rotatebox{90}{$\kappa_{L, \theta_{L}, \theta}(\cdot, x_{L-1};y)$}};
	\node[minimum height=1cm,fill=green!8] (out1) [right of=kL] {\rotatebox{90}{\dots}};
	\node[draw, minimum height=1cm, node distance=1.2cm,fill=green!8] (out) [below of=out1] {\rotatebox{0}{$x_{L0}$}};
	\node[draw, minimum height=1cm, node distance=1.2cm,fill=green!8] (out2) [above of=out1] {\rotatebox{0}{$x_{L}$}};

	\draw [->,out=30,in=210,relative]  (i) to (k0);
	\draw [->,out=30,in=210,relative]  (k0) to node [above, minimum height=0.5cm, fill = none] {\rotatebox{0}{$\hat x_0$}} (kdot);
	\draw [->,out=30,in=210,relative]  (kdot) to   (kl);
	\draw [->,out=30,in=210,relative] (kl) to node [above, minimum height=0.5cm, fill=none] {\rotatebox{0}{$\hat x_\ell$}} (klp1);
	\draw [->,out=30,in=210,relative] (klp1) to  (kdot2);
	\draw [->,out=30,in=210,relative] (kdot2) to (kL);
	\draw [->,out=30,in=210,relative] (kL) to node [above, minimum height=0.5cm, fill=none] {\rotatebox{0}{$x_L$}} (out1);
	
	\coordinate (L0_top) at ($(kdot.north east) + (0.375, 0)$);
	\coordinate (L0_bottom) at ($(kdot.south east) + (0.375, 0)$);
	\node[below=0cm, anchor=south, minimum height=0cm,fill=green!8] (l0) at (L0_top) {\rotatebox{0}{$L_0$}};
	
	\draw [loosely dashed] (L0_top) -- (L0_bottom);
	
\coordinate (above_start) at ($(k0.north) + (0, 0.6)$);
	\coordinate (below_start) at ($(kL.south) + (0, -0.6)$);
	\draw [->] (i) |- (above_start) -| (kl);
	\draw [->] (i) |- (above_start) -| (kdot);
	\draw [->] (i) |- (above_start) -| (klp1);
	\draw [->] (i) |- (above_start) -| (kdot2);
	\draw [->] (i) |- (above_start) -| (kL);
	
	\draw [->] (L0_bottom) |- (below_start) -| (out);
	\draw [->] (kl) |- (below_start) -| (out);
	\draw [->] (klp1) |- (below_start) -| (out);
	\draw [->] (kdot2) |- (below_start) -| (out);
	
	\begin{scope}[show background rectangle, background rectangle/.style = {fill = gray!0},, local bounding box=scopekey]
	\draw [->,fill=none] (1,-3.5) -- node  [left,fill=none, minimum height = 0pt, anchor = east, node distance =2cm, outer sep =32pt] {Key:} node [above,fill=none, minimum height=0pt] {Deterministic} (2,-3.5);
	\draw [->,out=30,in=210,relative,fill=none] (3.5,-3.5) to node [above,fill=none,minimum height=0pt] {Stochastic} (4.5,-3.5);
	
	\node[fit=(scopekey), draw=black, inner sep=0pt,fill=none,minimum height=0pt] {};
	\end{scope}
\end{tikzpicture}
	}
	\caption{} 
	\label{fig:intro/unfolded_mcmc_diagram}
	\end{subfigure}
    
	\caption{\textbf{(a):} Illustration of unfolded optimization architecture. The parameters $\theta, \vartheta_{0}, \dots, \vartheta_{L}$ are trained end-to-end such that the output samples from layers $\hat x_L$ is a close approximation of $\mathbb{E}[x|y]$. \textbf{(b):} Illustration of the proposed unfolded MCMC architecture. Parameters $\theta, \vartheta_{0}, \dots, \vartheta_{L}$ are trained end-to-end such that the output samples from layers $\ell\ge L_0$ closely resemble $p(x|y)$ over a training dataset.}
	\label{fig:intro/diagram}
\end{figure}

\subsection{Data-Driven Priors}
Recent developments in machine learning propose several approaches for modelling data-driven priors. In this work, we consider both plug-and-play denoisers and score based techniques, which are summarised below.
Plug-and-play (PnP) techniques relate the score $\nabla \log p_\theta$ to a minimum-mean-square-error (MMSE) denoiser. Introducing $\varepsilon\ll 1$, Tweedie's identity \cite{efron2011tweedie}  approximates the score $\nabla_x \log p(x) \approx  (D_\varepsilon(x) - x)/\varepsilon$, where $D_\varepsilon(z) = \mathbb{E}_{\rvx \sim p(x)}[\rvx|\{\rvz = z\}]$ is a minimum-mean-square error (MMSE) denoiser of $\rvz \sim \mathcal N(\rvx, \varepsilon I)$. For structured data-distributions, $D_\varepsilon$ can be well-approximated by a deep networks $D_{\varepsilon, \theta}$ such as a convex ridge-regulariser \cite{goujon2023neural},  CNN \cite{terris2020CNN, zhang2017beyond} or DRUNET \cite{zhang2021plug}. The weights $\theta$ are trained to minimise the mean-square loss over data $\mathcal X$
\begin{linenomath*}\begin{equation}
\label{eqn:meansquare_pnp}
\theta^\star = \argmin_\theta \E_{\rvx\sim p(x), \rvw\sim \mathcal N(0,1)}\Big[\big\|\rvx - D_{\varepsilon, \theta}(\rvx + \sqrt{\varepsilon}\rvw) \big\|^2\Big].
\end{equation}\end{linenomath*}
This provides an implicit representation of $p_\theta$ by estimating the score, which can be combined within gradient-based optimization/MCMC pipelines.

\subsection{Optimization Schemes and Deep Unfolding}
Optimization methods can be used to seek a maximiser $\hat x_\text{MAP}$ of $\pi_\theta(x|y)$. A broad class of explicit numerical optimization methods can be expressed through a recurrence of the form
\begin{linenomath*}\begin{equation} 
	\label{eqn:intro/optim}
	\hat x_{\ell+1} = G_{\ell, \vartheta_\ell, \theta}(\hat x_\ell;y), \qquad \ell\in\mathbb{N},
\end{equation}\end{linenomath*}
with transitions $G_{\ell, \vartheta_\ell, \theta}:\xspace\times \yspace\to \xspace$ at iteration $\ell>0$ and an initialization $G_{0, \vartheta_0, \theta}:\yspace\to\xspace$ for $\ell=0$. The transitions $G_{\ell, \vartheta_\ell, \theta}$ often arise naturally as a discretisation of the gradient flow $x^{\prime}(t) = -\nabla_x(f_y(x) + g_\theta(x))$, with hyper-parameters $\vartheta_\ell$ at iteration $\ell$. Methods of the form \eqref{eqn:intro/optim}  include (accelerated) gradient descent \cite{nesterov2003introductory} and proximal algorithms \cite{daubechies2004iterative} along with plug-and-play architectures which leverage an pre-trained MMSE denoiser to approximate the score $\nabla_x\log p_\theta(x)$ \cite{hurault2021gradient, hurault2022proximal}.

Unfolded optimization \cite{chen2022l2o} refers to a class of methods whereby an iterative scheme \eqref{eqn:intro/optim}
\begin{linenomath*}\begin{equation} 
	\label{eqn:intro/optimization}
	\mathcal G_{\Theta}(\cdot) 
	=  
	G_{L,\vartheta_{L}, \theta}
	\circ \cdots \circ
	G_{0,\vartheta_{0}, \theta}(\cdot)  
\end{equation}\end{linenomath*}
is viewed as a $L+1$-layer  neural-network $\mathcal G_{\Theta}:\yspace\to\xspace$, with trainable weights $\Theta = [\vartheta_0, \dots, \vartheta_L, \theta]$ \cite{gregor2010learning}.  During an offline training phase, optimal weights $\widehat\Theta$ are computed by minimising a suitable objective function. For example, the mean-square loss
\begin{linenomath*}\begin{equation}
	\label{eqn:intro/optim_objective}
	\mathcal L_\text{opt}(\Theta) = \E_{\rvx,\rvy\sim p(x,y)} \Big[\underbrace{\| \rvx - \mathcal G_{\Theta}(\rvy) \|^2}_{l_\text{opt}(\rvx, \rvy)} \Big],
\end{equation}\end{linenomath*}
which is minimised over  all $p_\mathbf{y}$-measurable random variables by the posterior mean of $\rvx$ given $y$. Thus, the optimal $\hat \Theta$ for the objective \eqref{eqn:intro/optim_objective} will learn the mapping $y\mapsto \E_{\rvx\sim p(x)}[\rvx|\rvy=y]$ for $y\sim p(y)$.
In practice, $p(x)$ is known only empirically and the loss \eqref{eqn:intro/optim_objective} is computed by a Monte Carlo estimate using mini-batches from the dataset $\mathcal D_{N_\text{data}} = \{(x^{(\star, n)}, y^{(n)})\}_{n=1}^{N_\text{data}}$.

An illustration of a typical unfolded optimization network is displayed in \Cref{fig:intro/unfolded_gd_diagram}. By designing a network architecture in this way, we obtain a model which has a degree of physical interpretability in contrast to standard feed-forward neural networks. From a Bayesian perspective, by jointly optimising the prior weights $\theta$ with  the hyper-parameters $\vartheta_\ell$ of the chosen algorithm, the unfolded model balances a delicate trade-off between descriptive prior information and fast algorithm convergence. Given a suitable amount of training data, on average for a new observation $(x,y)\sim p_{\rvx,\rvy}$ the unfolded network \eqref{eqn:intro/optimization} with optimal weights $\widehat \Theta$ will obtain a lower error in the objective $l_\text{opt}(x, y)$ after $L$ gradient steps.  The idea of unfolding iterative procedures in this manner was first used to improve convergence of the iterative soft-thresholding algorithm for problems in sparse coding \cite{daubechies2004iterative, gregor2010learning}. This concept has since been applied to construct efficient optimization solutions in a variety of settings \cite{chen2022l2o} including computational imaging \cite{Monga2021, zhao2024deep, sanghvi2022photon}. Unfolded PnP algorithms for challenging problems in computational imaging can be found in \cite{Monga2021,sanghvi2022photon}.

\subsection{Markov Chain Monte Carlo}
\label{sec:background/mcmc}
MCMC methods generate posterior samples recursively by iteratively sampling from a Markov transition kernel 
\begin{linenomath*}\begin{equation}
\label{eqn:intro/kernel_transition}
\begin{aligned}
	\hat x_{\ell+1}&\sim \kappa_{\ell, \vartheta_\ell, \theta}(\cdot, \hat x_\ell;y),\qquad \ell\in\mathbb{N},
\end{aligned}
\end{equation}\end{linenomath*}
with the convention that $\hat x_0\sim\kappa_{0,\vartheta_0,\theta}$ such that $\kappa_{0, \vartheta_0, \theta}$ places an initial distribution for the chain. We assume throughout that for all $(x,y)\in \xspace\times\yspace$ the kernel $\kappa_{\ell, \vartheta_\ell, \theta}(\cdot;x,y)$ can be sampled with bounded computational cost. Allowing for a burn-in period $L_0\ge 0$, samples  $\{\hat x_\ell\}_{\ell\ge L_0}$ can be viewed as approximate (dependent) samples from $\pi_\theta(x|y)$. Such techniques can often be interpreted as the discretisation of a gradient flow on the space of measures \cite{wibisono2018sampling} through discretization of the Langevin diffusion
\begin{linenomath*}\begin{equation}
	\label{eqn:overdamped_langevin}
	\text{d}\rvx_t = -\nabla_x(f_y(\rvx_t) + g_\theta (\rvx_t))\text{d}t + \sqrt{2}\text{d}\rvw_t,
\end{equation}\end{linenomath*}
where $(\rvw_t)_{t\ge 0}$ is an $\xspace$-valued Brownian motion. Examples include the unadjusted Langevin Algorithm \cite{durmus2017nonasymptotic}, which is based on an Euler-Maruyama (E-M) discretisation of \eqref{eqn:overdamped_langevin}, along with proximal extensions \cite{durmus2018efficient,pereyra2016proximal}. Kernels of the form \eqref{eqn:intro/kernel_transition} also include splitting MCMC algorithms, for instance the split-Gibbs sampler \cite{Vono2019,Vargas-Mieles2022}. The inclusion of plug-and-play priors within the above MCMC kernels has been studied extensively in \cite{laumont2022bayesian,coeurdoux2023plug}.

Related to PnP-Langevin algorithms, score-based diffusion models (SBMs) \cite{Song2020ScoreBasedGM,Song2019} construct a generative model to approximately sample $p(x)$ by reversing the forward process 
\begin{linenomath*}\begin{equation}
	\label{eqn:background/forward_diffusion}
	\text{d}\rvx_t = -\beta_t\rvx_t\text{d}t + \sqrt{2\beta_t}\text{d}\rvw_t
\end{equation}\end{linenomath*} 
transporting $p(x)$ at $t=0$ to $\mathcal N(0,\id)$ as $t\to\infty$.  It follows that $\rvx_t$ is Gaussian with mean $\mu_t=\rvx_0\exp(-\int_0^t \beta_s\text{d}s)$ and noise level $\sigma_t^2$. The reversed process takes the form 
\begin{linenomath*}\begin{equation} 
	\label{eqn:score_based_diffusion}
	\text{d}{\rvx_t} = -(\beta_t{\rvx_t}/2 + \beta_t \nabla_x\log p_{t}({\rvx_t}))\text{d}t + \sqrt{\beta_t}\text{d}{\rvw_t},
\end{equation}\end{linenomath*} 
where we introduce the marginal $p_t(x) = \int_{\xspace} p_{0t}(x_t|x_0) p(x_0)\text{d}x_0$. The score $\nabla_x\log p_{t}({\rvx_t})$ can be approximated through Tweedie's identity by training an MMSE denoiser $D_{t, \theta}$ using a score-matching loss based on \eqref{eqn:meansquare_pnp} to learn the noise added to $\rvx_t$. The reversed diffusion process can then be sampled by E-M discretisation \cite{ho2020denoising} or implicit (DDIM) steps \cite{song2021denoising}. More recently, consistency models \cite{song23consistency} fine-tune $D_{t,\theta}$ to allow accurate sampling within a vastly reduced number of iterations.  

By replacing ${\nabla_x\log p_{t}({\rvx_t})}$ with the conditional score defined through 
\begin{linenomath*}\begin{equation}
	\label{eqn:background/cond_score}
	{\nabla_x\log p_{t}({\rvx_t}|y) = \nabla_x \log\int_{\xspace} p_{0t}(x_t|x_0) p(x_0|y)\text{d}x_0}
\end{equation}\end{linenomath*}
one can guide the reversed diffusion process to approximately sample $p(x|y)$ at time $t=0$.  In \cite{zhu2023denoising,chung2023diffusion}, this idea is used to construct zero-shot Bayesian sampling algorithms. Similar methodology is employed in \cite{LiuI2SB} using a different reverse diffusion process. Note that in general, the integral in \eqref{eqn:background/cond_score} is intractable and the conditional score is typically approximated analytically. For example, in \cite{chung2023diffusion} the authors impose that $p_{t0}(x_0|x_t) = \delta_{\hat\rvx_0}$, where $\hat \rvx_0 = \E[\rvx_0|\rvx_t]$. Recent methods \cite{garber2024zero, spagnoletti2025LATINO} propose related techniques using pre-trained consistency models to generate accurate posterior samples in a small number of diffusion steps. In \cite{mbakam2024}, the authors use a small number of DDIM steps to act as a plug-and-play prior for the Langevin diffusion.

For more accurate performance on a determined class of likelihood models, several works have considered fine-tuning the pre-trained score model over a training set $\mathcal D$ of $(x^{\star}, y)$ pairs to model the conditional score for a determined class of likelihood models instead. In \cite{zhaoCosign}, the authors approximate $\nabla_x\log p_{t}({\rvx_t}|y)$ by fine-tuning a pre-trained score $D_{t, \theta}$ using a control-net \cite{zhang2023controlnet} taking $y$ as input. The resulting model is thus tailored to fit specific likelihood models. A similar approach is taken in \cite{mbakam2025learning}, which unfolds the stochastic gradient LATINO algorithm,  based on the Langevin diffusion, \Cref{alg:latino} to sample from $p_0(x)$ given both $\rvx_t$ and $y$. We unfold the LATINO kernel in \Cref{sec:numerics/RI} for a radio interferometry problem.

\subsection{Deep Generative Modelling}
Generative machine-learning architectures model $p(x)$ as the push-forward of a latent measure $p_\rvz$ by a deep neural-network `generator' $G_\theta$. By choosing an easy to sample latent measure, for example $p_\rvz \overset{\text{d}}{=} \mathcal N(0,\id_{d_z})$, samples from the distribution $p_\theta = (G_\theta)_\#p_\rvz$ are straightforward to obtain. Examples include variational auto-encoders (VAE) \cite{kingma2013auto}, generative adversarial networks \cite{goodfellow2014generative} and normalising flows \cite{hagemann2023generalized,Hagemann22normalizing,hagemann2024posterior}. VAE's train an encoder-decoder pair to encode samples from $p(x)$ to a latent (Gaussian measure) $p_\rvz$ while jointly learning to decode latent variables $\rvz\sim p_\rvz$ to recover an approximation of the prior $p(x)$. Weights in the encoder and decoder are trained to optimize an evidence lower bound based on the Kullbach-Leibler divergence. An alternative approach for generative modelling includes Generative Adversarial Networks (GANs) \cite{goodfellow2014generative}, discussed in \Cref{sec:background/cgan}. 

Such techniques are commonly used for Bayesian inversion, by letting $G_\theta = G_\theta(\cdot, y)$ depend on the measurements $y$. Weights $\theta$ are then optimized such that $G_\theta(\cdot, y)_\# p_\rvz \approx p(x|y)$ for $y\sim\rvy$. Examples include conditional VAE \cite{zhang2021conditional} and conditional GANS, discussed below.

\subsection{Conditional Generative Adversarial Networks}
\label{sec:background/cgan}
GANs \cite{goodfellow2014generative} refer to generative networks trained end-to-end through joint optimization with an adversarial discriminator. A GAN targetting $p(x)$ can be thought of as a function $G_\theta:\rset^{d_l}\to \xspace$ such that for an $\rset^{d_l}$-valued $\rvz\sim p_\text{latent}$, $G_\theta(\rvz)$ is approximately a sample from $p(x)$. Wasserstein GANs \cite{arjovsky2017wasserstein} learn $\theta$ through a loss function related to the Wasserstein-1 loss. Letting $(G_\theta)_\# p_\text{latent}$ denote the push-forward measure of $p_\text{latent}$ by $G_\theta$, we have \begin{linenomath*}\begin{equation}
\label{eqn:background/w1}
	\begin{aligned}
		\mathcal W_1(p_\rvx, (G_\theta)_\# p_\text{latent}) &= \sup_{D \in \mathcal C_1}  \E_{\rvx\sim p(x)}[D(\rvx)] - \E_{\rvz\sim p(z)}[D(G_\theta(\rvz))]\\
		&\approx \sup_{\phi \in C_1^\phi}  \E_{\rvx\sim p(x)}[D_\phi(\rvx)] - \E_{\rvz\sim p_\text{latent}}[D_\phi(G_\theta(\rvz))],
	\end{aligned}	
\end{equation}\end{linenomath*}
where $\mathcal C_1$ is the space of 1-Lipschitz functions from $\rset^{d_l}\to \xspace$, and $C_1^\phi$ is the set of weights $\{\phi: D_\phi\in\mathcal C_1\}$ for a chosen parametrization $\{D_\phi:\phi\in \mathcal C_1^\phi\}\subset \mathcal C_1$. Here, $D_\phi$ is a discriminator, separating true samples of $p_\rvx$ from artificial data sampled through $G_\theta$. The weights $\theta$ and  $\phi$ are trained by jointly minimising \eqref{eqn:background/w1} for $\theta$, while maximising for $\phi$ over finite batches of samples from $p(x)$. The constraint $\sup_{\phi \in C_1^\phi}$ is typically relaxed through imposing a gradient penalty term, which acts as a Lagrangian \cite{gulrajani2017improved}. 

Conditional GANs \cite{isola2017image,adler2018, bendel2023regularized} extend this methodology to train a conditional generator ${G_{\theta}:\rset^{d_l}\times\yspace\to \xspace}$, embedding the posterior distribution $p(x|y)$, given both the latent variable $\rvz$ and $y$ as input. Letting $\rvx_\theta^y \sim G_\theta(\cdot, y)_\# p_\text{latent}$, the Wasserstein loss function targeting the joint-distribution becomes
\begin{linenomath*}\begin{equation}
		\label{eqn:background/w1_cgan}
	\mathcal W_1(p_{\rvx, \rvy}, p_{\rvx_\theta^\rvy, \rvy}) \approx \sup_{\phi\in\mathcal C_1^\phi} \E_{\rvx, \rvy \sim p_{\rvx, \rvy}}[D_\phi(\rvx, \rvy)] - \E_{\rvz\sim p_\text{latent}, \rvy\sim p_\rvy}[D_\phi(G_\theta(\rvz, \rvy), \rvy)],
\end{equation}\end{linenomath*}
where the discriminator $D_\phi:\rset^{d_l}\times\yspace\to\xspace$ encodes additionally the observation $\rvy$ as input. The weights $\theta$ and $\phi$ for the conditional GAN can be again solved through a min/max optimization process with a Lagrangian to promote the constraint on $\phi$. Training solely on a conditional adversarial objective in this manner typically leads to mode collapse among posterior samples \cite{bendel2023regularized}. As a result, it is beneficial to add extra regularisation terms promoting data consistency.  In \cite{bendel2023regularized}, an $\mathcal L_1$-regularisation term is proposed  based on the $P$-sample posterior mean with a carefully-tuned standard deviation reward of the form
\begin{linenomath*}\begin{equation}
	\label{eqn:background/condgan_l1_reg}
	\begin{aligned}
	\mathcal L_{1,\text{SD}}(\theta) = &\E\left[
		\left\|
			\rvx - \frac{1}{P}\sum_{i=1}^P G_\theta(\rvz^{(i)}, \rvy)
		\right\|_1
	\right]\\
	&- w_\text{SD}
	\sum_{i=1}^P
	\E\left[
	\left\|
	G_\theta(\rvz^{(i)}, \rvy) - \frac{1}{P}\sum_{i=1}^P G_\theta(\rvz^{(i)}, \rvy)
	\right\|_1
	\right].
	\end{aligned}
\end{equation}\end{linenomath*}
For $x, y\sim p_{\rvx, \rvy}$ the first component  promotes that the mean of $(G_\theta(\cdot, y))_\# p_\text{latent}$ lies close to $x$. The latter component is a standard deviation reward promoting diverse samples. During the training phase, the weight $w_\text{SD}$ is carefully tuned online using a moment-matching procedure during frequent validation phases to promote $\theta$ with an accurate standard deviation. This fine-tuning of $w_\text{SD}$ is performed using the observation that if $(G_\theta(\cdot, y))_\#p_\text{latent} = p(x|y)$, then 
\begin{linenomath*}\begin{equation}
	\label{eqn:background/std_update}
	\frac{\mathbb E[\|\rvx - G_\theta(\rvz, \rvy) \|^2]}{\mathbb E[\|\rvx - {N_\text{val}}^{-1}\sum_{i=0}^{N_\text{val}}\mathcal G_\theta(\rvz^{(i)}, \rvy) \|^2]} = 2\frac{N_\text{val}+1}{N_\text{val}}.
\end{equation}\end{linenomath*}
Empirically,  the left hand side of the above term with $\theta$ trained with penalty factor $w_\text{SD}$ is found to a monotone in $w_\text{SD}$. A Robbins-Monro step is thus employed during each validation phase to find the root $w_\text{SD}$ which satisfies the above inequality.  
In \Cref{sec:method/training}, we leverage this approach to learn weights for efficient posterior sampling through an unfolded Markov kernel  \eqref{eqn:unfolded_mcmc}.

\section{Proposed Methodology}
\label{sec:methodology}
\subsection{Architecture}
Inspired by deep unfolding of iterative optimization schemes, we propose a novel framework for unfolding MCMC chains to obtain efficient sampling of posterior data. Specifically, we view samples from the distribution
\begin{linenomath*}\begin{equation}
	\label{eqn:unfolded_mcmc}
	\mathcal K_{\Theta,L}(\cdot, y) = (\kappa_{L, \vartheta_L, \theta} \circ \cdots \circ \kappa_{0, \vartheta_0, \theta})(\cdot, y)
\end{equation}\end{linenomath*}
as outputs from a conditional generative network targeting the distribution $p(x|y)$ with trainable weights $\Theta = [\vartheta_0,\dots, \vartheta_L, \theta]$. In a crucial difference from unfolded optimisation networks, we promote weights $\Theta$ which provide an accurate representation of the entire posterior distribution. In particular, we wish to train $\Theta$ to ensure $\mathcal K_{\Theta, L}(x, y)$ is a good approximation of the posterior $p(x|y)$ given a new sample $y\sim p_\rvy$. 
We sample $\mathcal K_{\Theta,L}(\cdot, y)$ by generating a Markov chain through the transitions \eqref{eqn:intro/kernel_transition}. Allowing for a suitable burn-in period $L_0\ge 0$, we can treat the output of each layer $\ell\ge L_0$ as an approximate sample from the underlying distribution. Specifically, for a given $y\in\yspace$, we then define the $\rset^{(L+1-L_0)\dx}$-valued random variable $\rvx_{L_0:L, \Theta}^y = [\rvx_{L_0, \Theta}^y, \dots,  \rvx_{L, \Theta}^y]$ such that $\rvx_{L_0, \Theta}^y \sim \mathcal K_{\Theta, L_0}(\cdot, y)$ and  $\rvx_{\ell+1, \Theta}^y\sim \kappa_{\ell, \vartheta_{\ell}, \theta}(\cdot, \rvx_{\ell+1, \Theta}^y; y)$ for $L_0\le\ell\le L-1$.

The proposed architecture is illustrated in \Cref{fig:intro/unfolded_mcmc_diagram}.  Leveraging the interpretability of each Markov transition \eqref{eqn:intro/kernel_transition}, the resulting model strikes a careful balance between the computational efficiency and complex embeddings of black-box conditional GANs and the modularity of MCMC techniques. Through aggregating dependent samples along trajectories of this truncated Markov Chain, a numerical study in \Cref{sec:numerics} shows the proposed unfolded MCMC method has a natural ability to produce reliable and efficient uncertainty and statistical analyses for new observation data.

\subsection{Regularised Adversarial Training}
\label{sec:method/training}
The proposed method adapts ideas from conditional GANs to fit the Markovian nature of the unfolded MCMC network. In summary: we apply conditional adversarial training to promote $\rvx_{\ell+1, \Theta}^y\sim\kappa_{\ell, \vartheta_{\ell}, \theta}$ to be approximately distributed like $p(x|y)$ for $L_0\le \ell\le L$. Following \cite{bendel2023regularized}, we add $\mathcal L_1$-regularisation to promote the sample mean 
\begin{linenomath*}\begin{equation*}
\overline{\rvx}_{L_0:L, \Theta}^y := \frac{1}{L-L_0+1}\sum_{\ell=L_0}^L \rvx_{\ell, \Theta}^y
\end{equation*}\end{linenomath*} 
to lie close to ground-truth data in mean-square error. For unfolded networks, the gradients of a loss function calculated only on the output $\hat x_L$ with respect to parameters $\vartheta_\ell$ with $\ell$ fixed are known to vanish as $L$ increases. By including the sample mean in the loss calculation, we abate the vanishing gradient problem prevalent in unfolded networks. 

The weights $\Theta$ are learned to minimise a training loss of the form \cite{bendel2023regularized}
\begin{linenomath*}\begin{equation}
    \label{eqn:full_loss_function}
    \mathcal L(\Theta) = \underbrace{\mathcal L_\text{adv}(\Theta; \phi)}_{\text{Adversarial Loss}} + \underbrace{w_1\mathcal L_1(\Theta)}_{\text{Data-consistency}} - \underbrace{w_{\text{SD}}\mathcal L_{\text{SD}}(\Theta)}_{\text{Sample diversity}}, 
\end{equation}\end{linenomath*}
where $w_1$ and $w_{\text{SD}}$ are positive regularisation weights and the learnable weights $\phi$ parametrise an adversarial discriminator as in \Cref{sec:background/cgan}. We consider each component of this loss separately in detail below. 

The adversarial loss $\mathcal L_\text{adv}$ is designed to move the measures $\mathcal K_{\ell, \Theta}$ close to $p(x|y)$ for $L_0\le \ell\le L$. We apply a conditional Wasserstein loss \eqref{eqn:background/w1_cgan} to randomly sampled layers $\rvl\sim \mathcal U(\{L_0,\dots,L\})$ of the unfolded MCMC network. In particular, we motivate $\mathcal L_\text{adv}$ through the approximation
\begin{linenomath*}\begin{equation}
	\label{eqn:intro/W1}
	\begin{aligned}
		&\sum_{\ell=L_0}^L \mathcal W_1\big(p_{\rvx, \rvy},\ p_{\rvx_{\ell, \Theta}^\rvy, \rvy}\big)\\ 
		&\propto \E_{
				\rvl\sim\mathcal U(\{L_0,\dots,L\})}\bigg[\mathcal W_1\big(p_{\rvx, \rvy},\ p_{\rvx_{\rvl, \Theta}^\rvy, \rvy}\big) \bigg]\\
		&= 
			\E_{\rvl\sim\mathcal U(\{L_0,\dots,L\})}
				\bigg[ 
					\sup_{D \in \mathcal C_1}
					\left\{
					\E_{\rvx, \rvy \sim p_{\rvx, \rvy}}[
						D(\rvx, \rvy)
					] 
					- 
					\E_{\rvx_{\rvl, \Theta}^\rvy, \rvy \sim p_{\rvx_{\rvl, \Theta}^\rvy, \rvy}}[
						D(\rvx_{\rvl, \Theta}^\rvy, \rvy)
					]
					\right\}
				\bigg]\\
		&\approx \E_{\rvl\sim\mathcal U(\{L_0,\dots,L\})}\left[
			\sup_{\phi \in C_1^\phi}
			\left\{
			 \E_{\rvx, \rvy \sim p_{\rvx, \rvy}}[D_\phi(\rvx, \rvy)] 
			 -
			 \E_{\rvx_{\rvl, \Theta}^\rvy, \rvy \sim p_{\rvx_{\rvl, \Theta}^\rvy, \rvy}}[
			 	D_\phi(\rvx_{\rvl, \Theta}^\rvy, \rvy)
		 	]
		 	\right\}
		 	\right]
	\end{aligned}
\end{equation}\end{linenomath*}
where we use the same terminology as \Cref{sec:background/cgan}. In particular, $D_\phi$ is a deep neural-network based discriminator with learnable weights $\phi$ tuned to solve the maximisation problem in \eqref{eqn:intro/W1}. For a given $\phi\in C_1^\phi$, we thus consider
\begin{linenomath*}\begin{equation*}
\mathcal L_\text{adv}(\Theta; \phi) = 
\E_{\rvl\sim\mathcal U(\{L_0,\dots,L\})}\left[
\E_{\rvx, \rvy \sim p_{\rvx, \rvy}}[D_\phi(\rvx, \rvy)] 
-
\E_{\rvx_{\rvl, \Theta}^\rvy, \rvy \sim p_{\rvx_{\rvl, \Theta}^\rvy, \rvy}}[
D_\phi(\rvx_{\rvl, \Theta}^\rvy, \rvy)
]
\right].
\end{equation*}\end{linenomath*}
Following standard practice for training GANs \cite{goodfellow2014generative, arjovsky2017wasserstein}, we learn $\phi$ jointly with $\Theta$. Introducing a penalty $\mathcal L_\text{GP}$ to promote the condition $\phi\in C_1^\phi$, $\phi$ is trained to minimize the loss
\begin{linenomath*}\begin{equation*}
	\mathcal L_\text{disc}(\phi;\Theta) = -\mathcal L_\text{adv}(\Theta;\phi) + \mathcal L_\text{GP}(\phi;\Theta).
\end{equation*}\end{linenomath*}
As in \cite{bendel2023regularized}, $\mathcal L_\text{GP}$ is a gradient penalty adapted from \cite{gulrajani2017improved} to discriminators trained on the joint distribution $p_{\rvx, \rvy}$
\begin{linenomath*}\begin{equation*}
	\mathcal L_\text{GP}(\phi;\Theta) = 
	\E_{\substack{\alpha\sim \mathcal U([0,1])\\
		\rvl\sim \mathcal U(\{L_0,\dots, L\})\\
		 \rvx,\rvy\sim p_{\rvx, \rvy}\\
		 \rvx_{\rvl, \Theta}^y\sim \mathcal K_{\rvl, \Theta}(\cdot;\rvy)}}
	 \left[
	 	\Big(\big\|\nabla_x D_\phi(\alpha\rvx + (1-\alpha)\rvx_{\rvl, \Theta}^\rvy, \rvy)\big\| - 1\Big)^2 \right],
\end{equation*}\end{linenomath*}
which promotes $D_\phi$ to be 1-Lipschitz on the space of feasible image pairs. In practice, both $\mathcal L_\text{adv}$ and $\mathcal L_\text{disc}$ are approximated using minibatches of data, with randomly sampled layers $\ell\sim \rvl$ among which we choose to discriminate samples. The weights $\theta$ are then learned using stochastic optimisation techniques. In \cite{stanczuk2021wasserstein}, it was shown that joint optimization of the objectives $\mathcal L_\text{adv}$ and $\mathcal L_\text{disc}$ using mini-batching leads to a heavily biased approximation of the Wasserstein distance. A Lipschitz-continuous discriminator was shown empirically to create an accurate solution landscape for the min-max game \eqref{eqn:intro/W1}. 

As discussed in \Cref{sec:background/cgan}, training solely on an adversarial objective typically leads to mode collapse among posterior samples. To avoid this, as a key development for our method we adapt the regularisation term \eqref{eqn:background/condgan_l1_reg} borrowing from the concept of skip connection components in deep neural networks to fit the Markovian nature of the unrolled measure \eqref{eqn:unfolded_mcmc}. The data consistency loss is defined as
\begin{linenomath*}\begin{equation*}
    \mathcal L_1(\Theta) = \mathbb E_{\rvx, \rvy\sim p_{\rvx, \rvy}}\Big[\big\|\rvx -  \overline{\rvx}_{L_0:L, \Theta}^\rvy  \big\|_1\Big],
\end{equation*}\end{linenomath*}
promoting the property that the (truncated) ergodic average of the unfolded MCMC chain should act as a good point estimate of the ground-truth data. To promote diverse samples along each MCMC trajectory, we further adapt the standard deviation reward to include the standard deviation between dependent MCMC samples
\begin{linenomath*}\begin{equation*}
    \mathcal L_\text{SD}(\Theta) = \mathbb E_{\rvx, \rvy \sim p_{\rvx,\rvy}}\bigg[\sum_{\ell=L_0}^L\Big\|\rvx_{\ell, \Theta}^\rvy\ - \overline{\rvx}_{L_0:L, \Theta}^\rvy\Big\|_1\bigg].
\end{equation*}\end{linenomath*}
As discussed in \Cref{sec:background/cgan}, the weight $w_\text{SD}$ has to be carefully tuned during training to prevent over-fitting through inducing instabilities in the MCMC chain. The rule \eqref{eqn:background/std_update} used to tune $w_\text{SD}$ in \cite{bendel2023regularized} relies on having independent samples from $\pi_\theta(x|y)$. Therefore, we adapt $w_\text{SD}$ using a variation of \eqref{eqn:background/std_update} which based on independent samples $\{\rvx_{L, \Theta}^{(y,n)}\}_{i=1}^{N_\text{val}}$ be from $\mathcal K_{L, \Theta}(\cdot;y)$ for $y\in\yspace$. That is, we therefore adapt $w_\text{SD}$ using a Robbins-Monro  update during a validation stage the term
\begin{linenomath*}\begin{equation*}
	\frac{\mathbb E_{\rvx, \rvy\sim p_{\rvx, \rvy}}[\|\rvx - \rvx_{L, \Theta}^\rvy \|^2]}{\mathbb E_{\rvx, \rvy\sim p_{\rvx, \rvy}}[\|\rvx - {N_\text{val}}^{-1}\sum_{i=0}^{N_\text{val}}\mathcal \rvx_{L, \Theta}^{(\rvy,n)} \|^2]} = 2\frac{N_\text{val}+1}{N_\text{val}},
\end{equation*}\end{linenomath*}
with $\Theta =\Theta(w_1)$ through \eqref{eqn:full_loss_function}. We find empirically that this term is monotone in $w_\text{SD}$ for $w_\text{SD}$ sufficiently small. To prevent instabilities arising in the MCMC architecture from promoting excessive variability between Markov iterates, we set $w_\text{SD} = 0$ whenever the error 
\begin{linenomath*}\begin{equation*}
\mathbb E_{\rvx, \rvy\sim p_{\rvx, \rvy}}\left[\big\|\rvx - \rvx_{L, \Theta}^{\rvy} \big\|^2\right]
\end{equation*}\end{linenomath*} 
is observed to rise above a user-defined threshold.

\subsection{Regularised Training with Perceptual Qualities}
\label{sec:method/perceptual}
Perceptual image quality metrics have recently proven effective for the distillation of diffusion models in imaging applications.  By promoting a sharper image quality than standard $\mathcal L_1$ or $\mathcal L_2$ loss functions, the learned perceptual image patch similarity metric (LPIPS) \cite{zhang2018unreasonable} has been used to distil diffusion models for prior \cite{song23consistency} and posterior \cite{zhaoCosign} sampling. For problems over distributions of natural images, we therefore extend the loss function \eqref{eqn:full_loss_function}  to
\begin{linenomath*}\begin{equation} \label{eqn:full_loss_fn_images}
    \mathcal L_\text{im}(\Theta) = \mathcal L(\Theta) + \omega_{\text{PS}}\mathbb E_{\rvx,\rvy\sim p_{\rvx, \rvy}}\Big[\text{LPIPS}\big(\rvx, \rvx_{L,\Theta}^\rvy\big)\Big],
\end{equation}\end{linenomath*}
where LPIPS is the learned perceptual image patch similarity metric \cite{zhang2018unreasonable}.

\section{Numerical Experiments}
\label{sec:numerics}
We benchmark unfolded MCMC networks as discussed in \Cref{sec:methodology} on two imaging inverse problems: a proof of concept experiment on image deblurring of MNIST digits (\Cref{sec:numerics/mnist}) allowing for extensive tests and accurate computation of comparison metrics due to the small size and relative simplicity of the prior distribution; and Radio interferometry on galaxy images (\Cref{sec:numerics/RI}) for a realistic benchmark within a target application area where fast, interpretable inference is challenging yet highly desirable. We unfold a different discretisation of the Langevin diffusion \eqref{eqn:overdamped_langevin} for each problem, illustrating the generality of the proposed approach.
\footnote{All results were computed using a HPC filesystem equipped with a Nvidia A40 (48GB) GPU and an AMD EPYC 7543 32-Core Processor. The code required to for these experiments uses the DeepInverse Pytorch imaging library \cite{tachella2025deepinverse} and can be found at \href{https://github.com/JSpence97/UMCMC}{https://github.com/JSpence97/UMCMC}.}

\subsection{Image Deblurring on MNIST}
\label{sec:numerics/mnist}
We consider a class of inverse problems trained on digits from the MNIST dataset \cite{lecun1998gradient}. For compatibility with existing deep convolutional neural-network architectures, we add $2\times2$ zero-padding to each digit so that $\dx = 32\times32$. The relatively small dimension of this problem allows for efficient training of several configurations and robust ablation comparisons.
 
\subsubsection{Task}
We consider a linear inverse problem of the form \eqref{eqn:linear_inverse_problem}, with $\op x = k*x$ representing convolution with a motion-blur kernel $k$. As remarked above, unfolded MCMC architectures are able to naturally embed model parameters. This allows the training of deep unfolded generative networks robust to small perturbations in the observation model, represented in this instance through the kernel $k$.  To demonstrate this, for each observation $y$ we sample the kernel $k$ randomly as a realisation of $\rvk\sim \mathcal{GP}(11, \lambda_\text{Matern} = 0.3, \sigma_\text{Matern}=0.25)$, representing a random 2-dimensional trajectory embedded on a grid size $11\times11$, simulated independently from $x$ and $y$ as a 2-dimensional trajectory from a Gaussian process with a Mat\'ern covariance function with length scale 0.3 and standard deviation $0.25$. This procedure for sampling $\rvk$ is described in \cite{schuler2016learning}. 
We assume that the realisation of $\rvk$ is known at inference time. Therefore, for each tuple $(x,y,k)$, the posterior \eqref{eqn:bayesian} becomes $\pi_\theta(x|y,\rvk=k)\propto p(y|x, \rvk=k)p_\theta(x)$. 

\subsubsection{Model}
To model the prior $p_\theta(x)$, we first introduce an auxiliary variable from a Laplace distribution $\rvz \sim p_{\lambda,\rvz}(z) \propto\exp(-\lambda\|z\|_1)$, depending on parameter $\lambda$ controlling the desired sparsity level.  We assume that the distribution of $\rvx$, given $\rvz=z$ is Gaussian $p_W(x|z) = \mathcal N(x;\textnormal{Sig}(Wz), \rho^2)$ where $W:\rset^{dz}\to\xspace$ is a specified linear operator and $\textnormal{Sig} = (1+\exp(-x))^{-1}$ denotes the sigmoid function, forcing the mean of $p_W(x|z)$ to lie within the unit hypercube where MNIST images are contained. Hence, for $\theta = (\lambda, W)$ we have $p_\theta(x) \propto \int_{\rset^{d_z}} p_{W}(x|z)p_{\lambda,\rvz}(z)\text{d}z$. Exact computation of $p_\theta(x)$ is precluded by this high-dimensional non-standard integral. Instead, we consider a splitting approach \cite{Vargas-Mieles2022}, which samples $\rvx$ by marginalising the joint distribution $\pi_\theta(x,z|y)\propto p(y|x)p_W(x|z)p_{\lambda,\rvz}(z)$.

Within this framework, we consider a split-Gibbs-sampler (SGS) asymptotically sampling $(\rvx_\ell, \rvz_\ell)$ from the joint distribution $p_{\rvx,\rvz}$ by iteratively sampling $\rvz_{\ell+1}\sim p(z| \rvx_\ell)$ and $\rvx_{\ell+1} \sim p(x | y, \rvz=\rvz_{\ell};\rho=\rho_\ell)$. Since $\rvx$ if a Gaussian random variable given $\rvz=\rvz_{\ell}$, $\rvx_{\ell+1}$ can be sampled exactly from a Gaussian distribution. The distribution $p(z| \rvx_\ell)$ is known only up to a normalizing constant and cannot feasibly be sampled exactly. Instead, we replace exact sampling of $p(z|\rvx_\ell)$ at each step with a single iteration of a Langevin  \eqref{eqn:overdamped_langevin} targeting the invariant measure $p(z| \rvx_\ell)$. The resulting sampler leads to the iterates \cite{Vargas-Mieles2022} 
\begin{linenomath*}\begin{equation}
	\label{eqn:numerics/gibbs}
	\begin{aligned}
		\rvz_{\ell+1} &= \rvz_\ell + \gamma_\ell\left(\nabla_z\log p(\rvx_\ell | \rvz = \rvz_\ell;\rho) + \nabla_z\log p_\rvz(\rvz_\ell)\right) + \sqrt{2\gamma_\ell}\zeta_\ell^z\\
		\rvx_{\ell+1} &\sim p(x | y, \rvz=\rvz_{\ell+1};\rho),
	\end{aligned}
\end{equation}\end{linenomath*}
where $\zeta_\ell^{z}$ are independent $\mathcal{N}(0,1)$ random variables for $0\le \ell\le L-1$.  Instead of forcing $\rho$ to be uniform between iterations, we allow for $\rho=\rho_\ell$ to depend on the iteration number $\ell$. This aligns with the stochastic approximation proximal gradient algorithm \cite[Algorithm 3]{Vargas-Mieles2022}, \cite{de2020maximum}. Such techniques numerically solve for the optimal value of $\rho$ for the data through an augmented stochastic approximation to find a root of $\nabla_\rho \log p(y|z;\rho)$.
Since the distribution of $\rvx$ given $\rvz = z$ is assumed Gaussian with mean $\textnormal{Sig}(Wz)$ and variance $\rho_\ell^2\id_{\dx}$, with $\rho=\rho_\ell$, we can write 
\begin{linenomath*}\begin{equation}
	\label{eqn:numerics/gibbs_l1}
	\begin{aligned}
	\rvz_{\ell+1}&= \rvz_{\ell} + \gamma_\ell\bigg((\nabla_z\textnormal{Sig}(W\rvz_\ell))^T\bigg(\frac{\rvx_\ell - \textnormal{Sig}(W\rvz_\ell)}{\rho_\ell^2}\bigg) + \frac{\text{ST}_{\lambda}(\rvz_\ell) - \rvz_\ell}{\lambda}\bigg) + \sqrt{2\gamma_\ell}\zeta_{\ell}^z\\ 
	\rvx_{\ell+1} &= \prox_{-\rho_\ell^2\log p(y | \cdot)}(\textnormal{Sig}(W\rvz_{\ell+1})) + \Big(\sigma^{-2}A^TA + \rho_\ell^{-2}\Big)^{-1/2}\zeta_\ell^x.
	\end{aligned}
\end{equation}\end{linenomath*}
The kernels $\kappa_{\ell, \vartheta_{\ell}, \theta}$ in \eqref{eqn:unfolded_mcmc} representing the update $\rvx_\ell\mapsto \rvx_{\ell+1}$ in \eqref{eqn:numerics/gibbs_l1} take weights $\theta = \{W, \lambda\}$ and $\vartheta_\ell = \{\gamma_\ell, \rho_\ell\}$. Since the Laplace prior enforces sparsity around feasible values of $z$, we initialize each chain with $\rvz_0 = \rvx_0 = 0$. That is, $\kappa_{0,\theta_0,\theta} = \delta_0$ is a Dirac measure. In what follows, we refer to the resulting unfolded MCMC chain as Unfolded-split Gibbs sampler (U-SGS).
We train U-SGS with kernels described implicitly through \eqref{eqn:numerics/gibbs_l1} using the training procedure discussed in \Cref{sec:numerics/mnist}. Independent models are trained for $L\in\{4,8,16,32,64\}$ iterations. For each model, we use a burn-in of size $L_0=L/4$.  For the adversarial loss, we use a discriminator $D_\phi$ based on \cite[Appendix I.1.3]{bendel2023regularized}, which is a convolutional network with 2 average-pooling layers which downsample by a factor of $2\times 2$ pixels, with batch-normalization and leaky ReLU activations. The final layer is fully-connected, mapping the output into $\rset$.

\subsubsection{Comparisons}
We compare the U-SGS architecture against a zero-shot MCMC method, along with an end-to-end neural network architecture trained to approximately sample $p(x|y,k)$. Each method is summarized below. 

For the zero-shot MCMC method, we use the same iterates in \eqref{eqn:numerics/gibbs}. Recall that the proposed U-SGS model relies crucially on tuning the weights $W$ in the Laplace prior for $\rvz$. Since this is not possible for a zero-shot method, we instead compare with a method which uses a pre-trained data-driven prior on $\rvz$ to promote a competitive comparison. Specifically, we parametrise a data-driven prior through a VAE encoder-decoder as in \cite{Holden2022}: given a latent variable $\rvz\sim p_\rvz^\text{latent} = \mathcal N(0, \id_{d_z})$, we use a pre-trained VAE encoder-decoder pair designed to encode $p_\rvx$ into $p_\rvz^\text{latent}$. Let $E_{\theta_e}^\text{VAE}:\xspace \to \rset^{d_z}$, $D_{\theta_d}^\text{VAE}:\rset^{d_z}\to \xspace$ denote the encoder and decoder, respectively. For this comparison, we use the prior model  $p_{\theta_d}(x) = (\text{D}_{\theta_d}^{\text{VAE}})_\# p_\rvz^\text{latent}$. We use the pre-trained encoder-decoder pair from \cite{Holden2022} which is a fully-connected network architecture with 2 hidden layers and a latent space of dimension $d_z=12$ for MNIST data. To sample from $\pi_{\theta_d}(x|y) =\propto p(y|x)p_{\theta_d}(x)$, we use the split Gibbs sampler \eqref{eqn:numerics/gibbs}, where $\rho_\ell\equiv \rho > 0$ is pre-trained on ground truth images $x$ alongside $\theta_d$ as part of the evidence lower bound approach \cite{Kingma2013AutoEncodingVB}. The gradients $\nabla_z\log p(\rvx_\ell | \rvz = \rvz_\ell;\rho=\rho_\ell)$ are computed using automatic differentiation. As is typical for zero-shot MCMC methods, we integrate \eqref{eqn:numerics/gibbs} for a large number of iterations $\ell\approx 10,000$ and approximate the ergodic mean as a point estimate of $x$ given $y$. We refer to this approach as VAE-SGS in what follows.

For the end-to-end neural network, we train a regularised conditional GAN (RCGAN) using the regularized loss \cite{bendel2023regularized} described by \eqref{eqn:background/w1_cgan}-\eqref{eqn:background/condgan_l1_reg}. For this comparison, we train an end-to-end generator $G_\theta(z,y)$, $z\sim \mathcal N(0, \id_{\dy})$, which has a U-Net architecture \cite{Ronneberger2015UNet} with two convolutional downsampling and two transpose convolutional upsampling layers with average pooling. The discriminator architecture is identical to the U-SGS model. While this comparison uses a similar training objective to U-SGS; the RCGAN uses a different generative architecture which is a conventional UNet that does not explicitly embed information regarding the Bayesian inverse problem.

In terms of architecture, VAE-SGS relies upon the same numerical integration scheme \eqref{eqn:numerics/gibbs} as U-SGS, but uses a different, zero-shot prior on $\rvx$. RCGAN generates samples from a single evaluation of a UNet, resulting in a more opaque sampler than U-SGS and VAE-SGS. However, like U-SGS, the RCGAN is trained in a few-shot manner from pairs $(x,y)\sim p_{\rvx,\rvy}$.

\subsubsection{Metrics}
We compare methods with a range of metrics quantifying both the accuracy of point estimates of $x^\star$ and the learned posterior distribution. The metrics are computed using a test set of 10,000 MNIST digits and kernels which we label $\mathcal D_\text{test} = \{(x_\text{test}^{(n)}, y_\text{text}^{(n)}, k_\text{test}^{(n)})\}_{n=1}^{10,000}$. To quantify data-fidelity, we compute the PSNR (in d.b.) for a single sample and for the ergodic sample average for Markovian methods and the 512-sample mean for the RCGAN. We compare perceptual sample qualities using the structural similarity index measure (SSIM) \cite{wang2004image} and learned perceptual image patch similarity (LPIPS) \cite{zhang2018unreasonable}.

We draw further comparisons using several metrics designed to compare the posterior approximation quality. In particular, we use the sliced Wasserstein (SW) distance \cite{bonneel2015sw} between $p_{\rvx, \rvy}$ and $p_{\rvx_\theta^{(\rvy)},\rvy}$ (see \Cref{app:metricscomparesw} for details). We further estimate the expected Wasserstein-2 distance, $\mathcal W_{2}^\text{Latent}$,  between the true and approximate posterior distributions pushed-forward to a latent Gaussian distribution by the pre-trained VAE encoder $E_{\theta_e}^\text{VAE}$. The metric is averaged over samples $y\sim \rvy$.  We remark that for MNIST data, $\mathcal W_{2}^\text{Latent}$ can be viewed as a more robust variant of the (conditional )Fr\'echet inception metric \cite{heusel2017gans,bendel2023regularized}. Further details of $\mathcal W_2^\text{Latent}$ can be found in \Cref{app:fid}. Since the latent embedding  $E_{\theta_e}^\text{VAE}$ was trained jointly with the prior distribution for VAE-SGS, it is expected that $\mathcal W_{2}^\text{Latent}$ will contain a small bias in favour of this method. However, we observe empirically in the following section and \Cref{tab:mnist_metrics} that any such benefits are minor. 

Finally, we compare distributions using the CMMD, which computes the maximum mean discrepancy between the push-forward of $p_\rvx$ and $p_{\rvx_\theta^\rvy}$ under a CLIP embedding \cite{jayasumana2024rethinking}.

\subsubsection{Results}
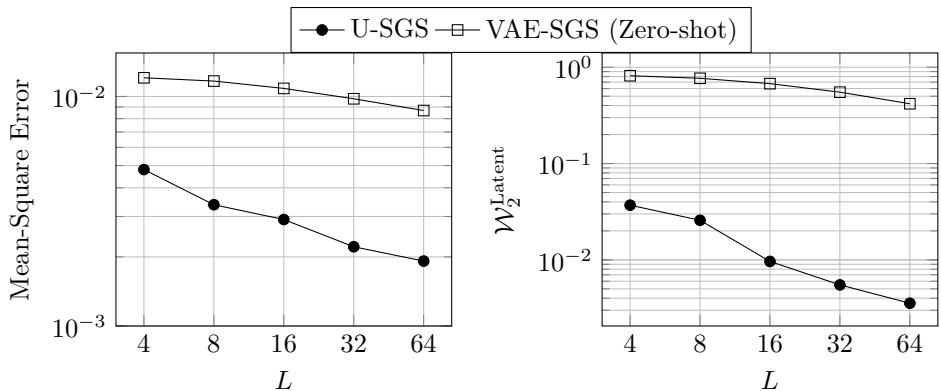
\begin{figure}
	\centering
	\begin{tikzpicture}
		\begin{axis}[
			name=ax1,
			xlabel=$L$,
			ylabel= Mean-Square Error,
			ymode=log,
			xmode = log,
			log basis x=2,
			log basis y=10,
			ymin=0.001,
			grid=both,
			legend style={at={(0.52,1)},anchor=south west, legend columns=-1},
			width=0.4\linewidth,
			xticklabels={,4,8,16,32,64}
			]
			\addplot[black,mark=*] table[x expr = \thisrow{L}, y expr = 10^(-\thisrow{Unrolled}/10), col sep = comma]{data/MNIST_mse.csv};
			\addlegendentry{U-SGS}
			\addplot[black,mark=square] table[x expr = \thisrow{L}, y expr = 10^(-\thisrow{VAE}/10), col sep = comma]{data/MNIST_mse.csv};
			\addlegendentry{VAE-SGS (Zero-shot)}
		\end{axis}
		\begin{axis}[
			at={(ax1.south east)},
			xshift=2cm,
			xlabel=$L$,
			ylabel= $\mathcal W_2^\text{Latent}$,
			ymode=log,
			xmode = log,
			log basis x=2,
			log basis y=10,
			grid=both,
			width=0.4\linewidth,
			xticklabels={,4,8,16,32,64}
			]
			\addplot[black,mark=*] table[x expr = \thisrow{L}, y expr = \thisrow{Unrolled_FID}, col sep = comma]{data/MNIST_mse.csv};
			\addplot[black,mark=square] table[x expr = \thisrow{L}, y expr = \thisrow{VAE_FID}, col sep = comma]{data/MNIST_mse.csv};
		\end{axis}
	\end{tikzpicture}
	\caption{Convergence comparison of U-SGS and conventional SGS on small truncated chains.}
	\label{fig:numerics/Conv_MNIST}
\end{figure}

Reconstruction metrics for each instance of our model and for both comparisons are shown in \Cref{tab:mnist_metrics}. After unfolding 8 iterations, our method is comparable to the RCGAN in terms of PSNR and is superior in terms of perceptual and distributional metrics. Increasing $L$ yields greater performance in our model at the expense of longer training and inference times. The model with $L=64$ iterations displays the best performance in all metrics, with an exception for SW which saturates (see \Cref{app:metricscomparesw} for an explanation).  The VAE-SGS method with 10,000 MCMC transitions exhibits comparable sample quality to the RCGAN but with less sample diversity, resulting in a worse PSNR for the posterior mean.  To illustrate the difference in convergence to the true posterior between the U-SGS and VAE-SGS over a small number of iterations, \Cref{fig:numerics/Conv_MNIST} compares the mean-square error and $\mathcal W_2^\text{Latent}$, averaged over the MNIST test set, for each method after $4\le L\le 64$ iterations.  Qualitative results are illustrated in \Cref{fig:qualitative_comparison_MNIST} for 3 typical inverse problems from the test set. For each method, we show compare the posterior mean (averaged over 512 posterior samples); the residual error, and the first three principal eigenvectors of the posterior covariance matrix, computed through a PCA expansion. We notice sharp sample quality for both our method and RCGAN. The RCGAN displays some artefacts on the middle and lower reconstructions of the characters `0' and `8'. This error is reflected in the principal eigen-directions for RCGAN, which are not contained in the manifold of MNIST data. This is expected since the GAN does not explicitly model the operator $\op$ and instead tries to directly learn the mapping $y \mapsto p_{\rvx|y}$ which causes difficulties in modelling the posterior over a range of kernels $k\sim\rvk$. In contrast, both U-SGS and VAE-SGS have a physical representation of the blur kernel through the likelihood. This allows both methods to adapt naturally to small perturbations in the blurring operation between inversions, exploring more insightful eigen-directions with the posterior samples.   
 VAE-SGS produces slightly blurry samples, indicating the method is concentrating around the parametrised posterior mean. The residual error and principal eigen-directions contain a lot of background noise, indicating the a significant amount of noise in the VAE-SGS posterior, perturbing samples away from the MNIST manifold. This noise could be diminished by reducing the step-size $\gamma$ in the VAE-SGS iterations, which comes at a greater inference cost. 
 
The above experiment shows unfolding an MCMC architecture effectively encodes $\op$ into the trained network which is beneficial when working with a range of similar forward operators. To highlight this property further, \Cref{app:mnist_ood} shows numerical results when comparing each of the above models on out-of-training-distribution operators $\op$. 

\begin{table}
	\centering
	\resizebox{\linewidth}{!}{
		\begin{tabular}{|c | c c c | c c c| c c|}
			\hline
			Method (NFEs) & \makecell{PSNR \\(sample)} & \makecell{PSNR \\ (mean)}  & LPIPS & \makecell{SW \\ ($\times 10^5$)}  & \makecell{$\mathcal W_2^\text{Latent}$ \\ ($\times 10^2$)} & \makecell{CMMD \\ ($\times 10^3$)} & \makecell{Time\\ (ms)} &\makecell{Size\\ (MB)} \\
			\hline
			\rowcolor{gray!12} U-SGS (4) & 23.77 & 24.96  & 0.007 & 9.53  & 3.70 & 12.70 &0.13 & \textbf{4.0} \\
			\rowcolor{gray!12}  U-SGS (8) & 24.95 & 26.29  & 0.006 & 8.49  & 2.58 & 3.91 & 0.24 & 4.0 \\
			\rowcolor{gray!12}  U-SGS (16) & 25.78 & 27.34  & 0.005 & 7.08  & 0.96 & 5.86 & 0.48 & 4.0 \\
			\rowcolor{gray!12}  U-SGS (32) & 26.93 & 28.12  & 0.003 & \textbf{6.72}  & 0.54 & 1.95 & 0.96 & 4.0 \\
			\rowcolor{gray!12}  U-SGS (64) & \textbf{27.59} & \textbf{28.46}  & \textbf{0.003} & 6.95  & \textbf{0.35} & \textbf{0.98} & 1.89 & 4.0 \\
			\hline
			RCGAN (1) & 23.28 & 26.69  & 0.009 & 9.21  & 9.08 & 21.48 & \textbf{0.10} & 11.83 \\
			\hline
			VAE-SGS ($10^4$) & 23.38 & 23.49  & 0.015 & 11.66  & 0.92 & 18.56 & 120 & 2.7 \\
			\hline
		\end{tabular}
	}
	\caption{Reconstruction metrics for the Random Motion deblurring problem. Results are averaged over the MNIST test dataset, using our unrolled Langevin model with $L=2^{i}$ iterations, $2\le i\le 6$. Comparisons are made with the rcGAN architecture \cite{bendel2023regularized} and a zero-shot split Gibbs sampler with a data-driven VAE prior (VAE-SGS). LPIPS is computed using a single approximate posterior sample from each model. For each observation $y$, PSNR results are computed for a single posterior sample and for the posterior mean, approximated using 512 Monte Carlo samples, with the exception of VAE-SGS for which the posterior mean is approximated by the ergodic average. The final columns show the time to generate a single posterior sample for each model, and the storage cost of model parameters. Metrics are scaled by the stated factors to assist readability.}
	\label{tab:mnist_metrics}
\end{table}

\newcommand\mywidthmnist{1.15}
\newcommand\mywidthmnistcm{1.15cm}
\begin{figure}
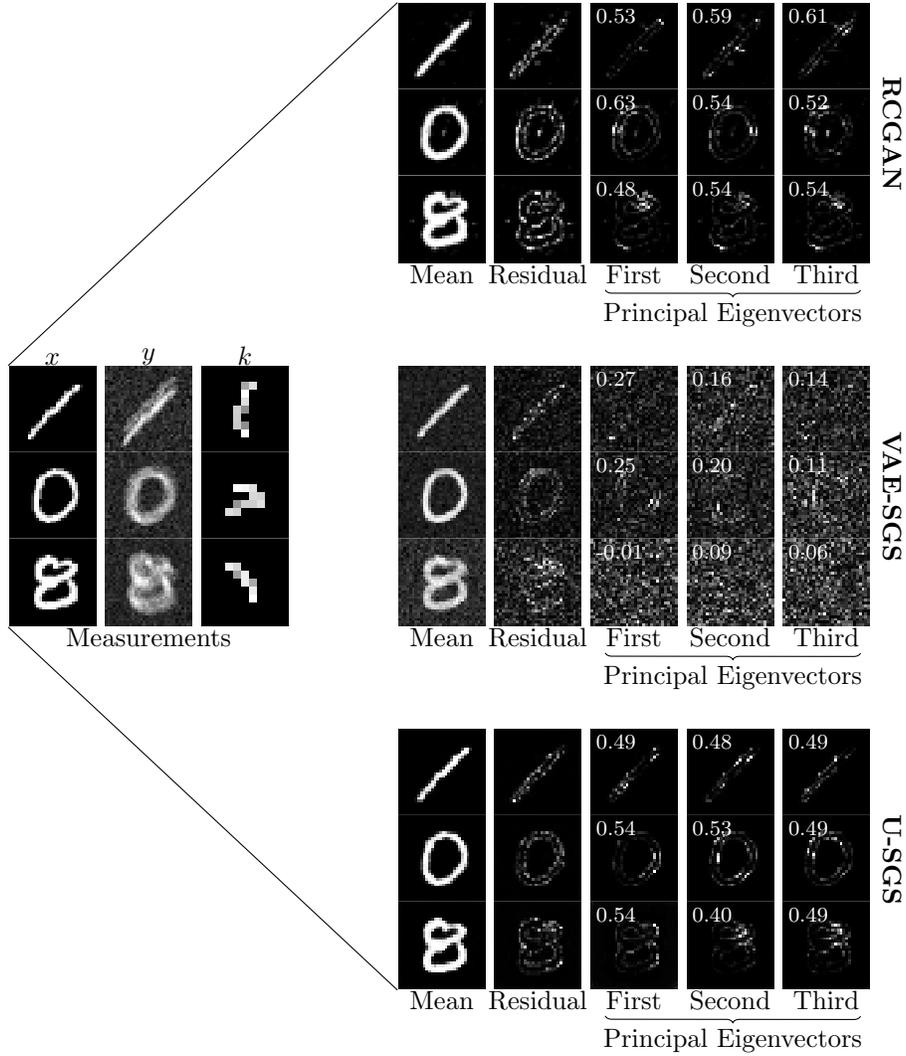

	\centering
	\begin{tikzpicture}[every node/.append style={rectangle, draw=white, thick, inner sep=0pt,}]

		\foreach \imno [count=\imcounter] in {11,13,15}{
			\node[inner sep = 0pt] (x\imcounter) at (-\mywidthmnist,-\imcounter * \mywidthmnist -3.65* \mywidthmnist) {\includegraphics[width=\mywidthmnistcm, height=\mywidthmnistcm]{figures/MNIST_Deblurring/visualizations/USGS/ground_truth_\imno.png}};
			\node[inner sep = 0pt, right=0.1cm] (y\imcounter) at (x\imcounter.east) {\includegraphics[width=\mywidthmnistcm, height=\mywidthmnistcm]{figures/MNIST_Deblurring/visualizations/USGS/measurements_\imno.png}};
			\node[inner sep = 0pt, right=0.1cm] (f\imcounter) at (y\imcounter.east) {\includegraphics[width=\mywidthmnistcm, height=\mywidthmnistcm] {figures/MNIST_Deblurring/visualizations/USGS/filter_\imno.png}};

			\foreach \methodno [count=\methodcounter] in {RCGAN_MNIST,VAE_MNIST,USGS}{

				\node[inner sep = 0pt] (m\imcounter\methodcounter) at (3.5 * \mywidthmnist,-4.2*\methodcounter* \mywidthmnist -\imcounter * \mywidthmnist + \mywidthmnist  + 3.75*\mywidthmnist) {\includegraphics[width=\mywidthmnistcm, height=\mywidthmnistcm]{figures/MNIST_Deblurring/visualizations/\methodno/avg_recon_\imno.png}};

				\node[inner sep = 0pt, anchor=west] (e\imcounter\methodcounter) at ($(m\imcounter\methodcounter.east) + (0.1,0)$) {\includegraphics[width=\mywidthmnistcm, height=\mywidthmnistcm]{figures/MNIST_Deblurring/visualizations/\methodno/err_\imno.png}};

				\node[inner sep = 0pt, anchor=west, right=0.1cm] (s\imcounter\methodcounter) at (e\imcounter\methodcounter.east) {\includegraphics[width=\mywidthmnistcm, height=\mywidthmnistcm]{figures/MNIST_Deblurring/visualizations/\methodno/principle_eigencomponent_0_\imno.png}};

				\node[inner sep = 2pt, anchor=north west,draw=none,] (corr\imcounter\methodcounter) at (s\imcounter\methodcounter.north west) {\textcolor{white}{\footnotesize \input{figures/MNIST_Deblurring/visualizations/\methodno/principle_eigencomponent_0_\imno.txt}}};

				\node[inner sep = 0pt, anchor=west, right=0.1cm] (s2\imcounter\methodcounter) at (s\imcounter\methodcounter.east) {\includegraphics[width=\mywidthmnistcm, height=\mywidthmnistcm]{figures/MNIST_Deblurring/visualizations/\methodno/principle_eigencomponent_1_\imno.png}};
				
				\node[inner sep = 2pt, anchor=north west,draw=none] (corr2\imcounter\methodcounter) at (s2\imcounter\methodcounter.north west) {\textcolor{white}{\footnotesize \input{figures/MNIST_Deblurring/visualizations/\methodno/principle_eigencomponent_1_\imno.txt}}};

				\node[inner sep = 0pt, anchor=west, right=0.1cm] (s3\imcounter\methodcounter) at (s2\imcounter\methodcounter.east) {\includegraphics[width=\mywidthmnistcm, height=\mywidthmnistcm]{figures/MNIST_Deblurring/visualizations/\methodno/principle_eigencomponent_2_\imno.png}};
				
				\node[inner sep = 2pt, anchor=north west,draw=none] (corr3\imcounter\methodcounter) at (s3\imcounter\methodcounter.north west) {\textcolor{white}{\footnotesize \input{figures/MNIST_Deblurring/visualizations/\methodno/principle_eigencomponent_2_\imno.txt}}};
				
				\ifthenelse{\imcounter>2}{
					\node[anchor=north, below] (m\methodcounter_title) at (m3\methodcounter.south) {Mean}; 
					\node[anchor=north, below] (e\methodcounter_title) at (e3\methodcounter.south) {Residual}; 
					\node[anchor=north, below] (s\methodcounter_title) at (s3\methodcounter.south) {First}; 
					\node[anchor=north, below] (s2\methodcounter_title) at (s23\methodcounter.south) {Second};
					\node[anchor=north, below] (s3\methodcounter_title) at (s33\methodcounter.south) {Third};
					\draw[decoration={brace,mirror,raise=0.5ex}, anchor=north,decorate] (s\methodcounter_title.south west) -- node [below=1.5ex,anchor=north] {Principal Eigenvectors} (s3\methodcounter_title.south east);
				}{}
				
			}

		}

		\node[anchor=south, above] (mlab1) at (x1.north) {$x$};
		\node[anchor=south,above] (mlab2) at (y1.north) {$y$};
		\node[anchor=south,above] (mlab3) at (f1.north) {$k$};
		
		\node[anchor=west,right=1ex] (slab1) at (s321.east) {\rotatebox{270}{\bf RCGAN}};
		\node[anchor=west,right = 1ex] (slab2) at (s323.east) {\rotatebox{270}{\bf U-SGS}};
		\node[anchor=west, right=1ex] (slab3) at (s322.east) {\rotatebox{270}{\bf VAE-SGS}};
		
\node[anchor=north, below] (ytitle) at (y3.south) {Measurements}; 
		
		\draw [] (x1.north west) -- (m11.north west);
		\draw [] (x3.south west) -- (m33.south west);

	\end{tikzpicture}
	\caption{Qualitative comparison for motion deblurring applied to the MNIST dataset. For each method, we display trained posterior samples from $p_\theta(x|y)$ for a range of three observations from the test dataset. In addition, the residual error and an estimate of the principal three eigen-directions from a PCA expansion are shown. For each eigenvector, the Pearson correlation to the residual is shown in the upper-left corner.}
	\label{fig:qualitative_comparison_MNIST}
\end{figure}

\subsection{Radio Interferometry}
\label{sec:numerics/RI}
To demonstrate the utility of unfolded an MCMC kernel in a more scientific experiment, we consider an application to radio interferometry (RI). We use $256\times256$ patches from radio telescope data.

\subsubsection{Task} We assume that the observation $y\in \yspace$ represents incomplete masked Fourier measurements from the interferometer with $\dy=\dx$. In order to reduce the computing time of the numerical experiments during training and testing phases, we exploit usual simplifications to the radio interferometric observation model \cite{dia2025iris}. In particular,  we  discretise the visibilities to avoid using a Non-Uniform FFT. We also adopt the coplanar baseline assumption such that all visibilities belong to a plane and we can handle a two-dimensional Fourier transform instead of its three-dimensional counterpart. Under this setup, we consider the observation model
\begin{linenomath*}\begin{equation}
\label{eqn:numerics/RI_task}
    y = \rvm\mathcal{F}P x^\star + \epsilon, \qquad \epsilon \sim \mathcal{CN}(0, \sigma_y^2\id ),
\end{equation}\end{linenomath*}
where $P\in \rset^{d\times d}$ is a diagonal matrix representing the pixelated primary beam of the antenna, $\mathcal F:\rset^d\to\rset^d$ is a 2D discrete FFT operator and the mask $\rvm = \mathcal F \rvk$ is the Fourier transform of a convolutional kernel $\rvk$ representing the forward observation model. Since input data is measured in the Fourier domain, we add complex Gaussian noise $\mathcal{CN}$ with isotropic variance $\sigma_y^2$. Adapting the setup in \cite{mars2025generative}, we generate a set  $K_\text{4h}$ of 10,200 kernels by simulating $uv$-coverages of the MeerKAT radio telescope with a 4 hour observation time. When generating each observation $y$, a random kernel $\rvk$ is sampled from a uniform measure over $K_\text{4h}$. 
Following \cite[Appendix A3]{dia2025iris}, we set $P=\id$ to simplify the covariance matrix $\op\op^T$. This facilitates comparison with the zero-shot SBM restoration technique in \cite{dia2025iris}.

We remark that in more realistic RI problems involving a Non-Uniform FFT, the forward operator becomes  costly to evaluate. Methods which maintain accurate performance while reducing the number of evaluations of the forward operator are therefore of great interest to the RI imaging community \cite{aghabiglou2024r2d2}. Compounding this fact with a requirement for accurate uncertainty estimates to inform quantitative downstream inferences about the ground truth data highlights the importance of fast and reliable statistical samplers for RI tasks. 

Due to requiring a large number of evaluations of the forward model, standard MCMC approaches are prohibitively expensive in realistic settings. An alternative approach is taken in \cite{liaudat2024scalable} use a convex approximation of the prior in order to exploit an approximation of the high posterior density region of log-concave models. However, the explicit convex prior has limited expressivity over the distribution of RI images, resulting in broad uncertainty intervals. In contrast, score-based models \cite{dia2025iris} leverage powerful prior information at the expense of significant compute time. 

\subsubsection{Data}
We train and benchmark our model on galaxy images from the Photometry and Rotation Curve Observations from Extragalactic Surveys (PROBES) dataset \cite{Stone2019PROBES,Stone2021PROBES}. This is a catalogue of over 2000 late-type galaxies exhibiting a diverse range of structural properties. During the unfolding phase, we leave 178 samples as hold-out images for benchmarking. 

\subsubsection{Model}
Using the terminology in \Cref{sec:background/mcmc}, we consider an implicit prior defined through a score-based denoiser $ \nabla \log p_{t,\varphi}(x)\approx (\mu_tD_{t,\varphi}(x) - x)/\sigma_t^2$. We load pre-trained score model weights $\varphi$ from \cite{dia2025iris}, which uses a deep residual U-Net architecture with attention layers \cite{song2019generative}. This model was trained using score-matching on the PROBES dataset for the variance-preserving diffusion process \eqref{eqn:score_based_diffusion}. The weights $\varphi$ have dimension of order $10^8$. To avoid unfolding a costly amount of weights, we follow the approach in \cite{mbakam2025learning} and represent $\varphi = \varphi_0 + \theta$, where $\varphi_0$ represents the frozen, pre-trained weights and $\theta$ represents a low-rank correction targetting the attention layers within $D_{t,\varphi}$. We unfold the LATINO architecture discussed in \Cref{alg:latino} with trainable weights $\theta$ and $\vartheta_\ell = (\gamma_\ell,t_\ell)$. To account for variability in the Lipschitz constant of $A$ between observations due to randomly sampling $\rvk\sim\mathcal U(K_\text{4h})$, we normalize each $\gamma_\ell$ by the largest component of $\rvm=\mathcal F \rvk$. LATINO can be expressed in the form \eqref{eqn:intro/kernel_transition} with kernel 
\begin{linenomath*}\begin{equation}
	\label{eqn:background/LATINO_kernel}
	\kappa_{\ell, \{t_\ell,\gamma_\ell\},\theta}(\cdot, \hat x_\ell;y) = (\prox_{\gamma_\ell f_y}\circ D_{t_\ell, \theta})_{\#}p_{0t_\ell}(\cdot|\hat x_\ell).
\end{equation}\end{linenomath*} 
For this experiment, we unfold $L=8$ iterations of \eqref{eqn:background/LATINO_kernel} using the methodology in \Cref{sec:method/training} with the additional perceptual loss discussed in \Cref{sec:method/perceptual}. We use an initial burn-in of size $L_0 = 2$.  We refer to the unfolded model as U-LATINO in what follows.

\begin{algorithm}
	\caption{LATINO \cite{spagnoletti2025LATINO}}
	\label{alg:latino}
	\begin{algorithmic}[1]
		\REQUIRE $y$, Score-based denoiser $D_{t, \theta}$, $x_0$, $\vartheta_\ell = \{t_\ell, \gamma_\ell\}$ for $0\le \ell\le L-1$.
		\STATE $\hat x_0\gets x_0$ \COMMENT{Initialization}
		\FOR{$\ell=0,\dots L-1$}
		\STATE Sample $\hat x_{\ell, t_\ell}\sim p_{0t_\ell}(x|\rvx_0 = x_\ell)$  \COMMENT{Forward SDE sample through \eqref{eqn:background/forward_diffusion}}
		\STATE $\hat z_{\ell+1}\gets D_{t_\ell, \theta}(x_{\ell,t_\ell})$ \COMMENT{Denoise step}
		\STATE $\hat x_{\ell+1}\gets \prox_{\gamma_\ell f_y}(z_{\ell+1})$ \COMMENT{Proximal data-fidelity step}
		\ENDFOR
		\RETURN $x_{L}$
	\end{algorithmic}	
\end{algorithm}

\subsubsection{Comparisons}
As a benchmark on few-shot RI reconstruction, we compare with RIGAN \cite{mars2025generative}. This model trains an RCGAN \cite{bendel2023regularized} for RI reconstruction.  For an accurate comparison on PROBES data, we re-train the generator for the task \eqref{eqn:numerics/RI_task} on PROBES data. For RIGAN, we use a U-Net generator architecture containing 4 each convolutional downsampling and upsampling layers. The RIGAN is trained on a normalized dataset of images with constant pixel mean and variance. For accurate comparisons, we rescale the RIGAN outputs to match the ground truth intensities. We make comparisons with two zero-shot methods: IRIS \cite{dia2025iris}, a state-of-the-art conditional SBM for RI image reconstruction; and LATINO \cite{spagnoletti2025LATINO} without unfolding. Both zero-shot models use the same pre-trained prior as U-LATINO. For IRIS, the conditional score  $\nabla_x\log p_{t}({\rvx_t}|y) = \nabla_x\log p_{t}(y|{\rvx_t}) + \nabla_x\log p_{t}({\rvx_t})$ is estimated by approximating $p_{t}(y|{\rvx_t})$ with a Gaussian convolution kernel under the assumption that $p_t({\rvx_t}) \approx p_0({\rvx_0})$. IRIS posterior samples are obtained by using E-M to approximately sample the conditional reversed diffusion discussed in \Cref{sec:background/mcmc}. In \cite{dia2025iris}, the authors use 4,000 E-M iterations to generate a single posterior sample. Due to computational limitations and to illustrate the dependence of IRIS on the number of iterations, we report IRIS results using both 64 and 1,000 E-M steps. With just 8 E-M steps, we observed numerical instabilities in IRIS reconstructions and we have thus chosen to omit IRIS results with the same computational cost per sample as U-LATINO. For LATINO, we use the same pre-trained SBM denoiser as IRIS and U-LATINO but without LoRA fine-tuning during an unfolding stage. We run \Cref{alg:latino} for $L=8$ iterations, with $\gamma_\ell =L_A/2$, $L_A$ representing the Lipschitz constant of $A$, and $t_\ell = 3(L+1-\ell)/4L$.  This comparison allows us to illustrate the performance gain by unfolding the LATINO MCMC kernel. For (U-)LATINO and IRIS experiments, a single NFE refers to one evaluation of the $D_{t,\varphi}$. For RIGAN, a single NFE refers to a single sample of the conditional GAN. 

\subsubsection{Metrics}
We report for each experiment the PSNR, LPIPS and SSIM metrics as in \Cref{sec:numerics/mnist}. To calibrate the PSNR, we infer the maximum intensity of each image from the ground truth data. In addition, we report the CFID using the procedure discussed in \Cref{app:fid}. For the purpose of this experiment, we use a VGG16 encoder to compute the FID embeddings and LPIPS scores. For this experiment, the PSNR was computed between the ground truth and the average of 8 posterior samples. All other metrics are computed using a single posterior sample for each observation.

\subsubsection{Results}
\begin{figure}
	\centering
	\begin{tikzpicture}[spy using outlines={circle, red, size=0.4*\mywidthcm, magnification=3}]
\node[inner sep = 1pt](x1) at (-0.4*\mywidth,-0.*\mywidth) {\includegraphics[width=\mywidthcm]{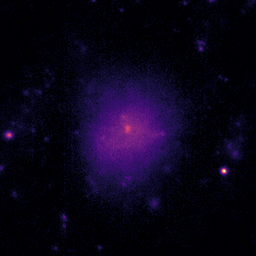}};
		\node[inner sep=0pt,anchor=north] (x1cmap) at (x1.south) {\includegraphics[]{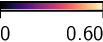}};
		\node[left] at (x1.west) {\rotatebox[]{90}{$x^\star$}};
		
		\node[inner sep=1pt] (y1) at (-0.4*\mywidth,-1.5*\mywidth) {\includegraphics[width=\mywidthcm]{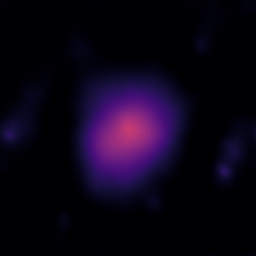}};
		\node[inner sep=0pt,anchor=north] (y1cmap) at (y1.south) {\includegraphics[]{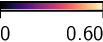}};
		\node[left] at (y1.west) {\rotatebox[]{90}{$A^\dagger y$}};
		
		\node[inner sep = 1pt](m1) at (-0.4*\mywidth,-3*\mywidth) {\includegraphics[width=\mywidthcm]{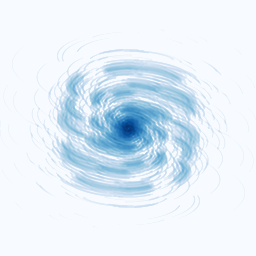}};
		\node[inner sep=0pt,anchor=north] (m1cmap) at (m1.south) {\includegraphics[]{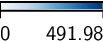}};
		\node[left] at (m1.west) {\rotatebox[]{90}{$\rvm$}};
\coordinate (spypoint) at ($(x1.north west) + (0.9*\mywidthcm, -0.65*\mywidthcm)$);
		\coordinate  (spypointviewer) at ($(x1.north west) + (\mywidthcm, -0.2*\mywidthcm)$);
		\spy  on (spypoint) in node [fill=white] at (spypointviewer);
		
		\coordinate (spypointy) at ($(y1.north west) + (0.9*\mywidthcm, -0.65*\mywidthcm)$);
		\coordinate  (spypointyviewer) at ($(y1.north west) + (\mywidthcm, -0.2*\mywidthcm)$);
		\spy  on (spypointy) in node [fill=white] at (spypointyviewer);

		\foreach \pth/\pthtitle/\pthNFEs/\imnopth [count=\pthct from 1] in {IRIS_probes_4h_1000_steps_img4/IRIS/1000/0, IRIS_probes_4h_64_steps_img4/IRIS/64/0, RCGAN_probes_4h/RIGAN/1/4, ULAT_probes_4h/U-LATINO/8/4, LAT_probes_4h/LATINO/8/4}{
			\begin{scope}[spy using outlines={circle, red, size=0.4*\mywidthcm, magnification=3}]
\node[inner sep=1pt](mean\pthct) at (0.5*\mywidth + \pthct *\mywidth + 0.45*\pthct,0) {\includegraphics[width=\mywidthcm]{figures/RI/visualizations/\pth/avg_recon_\imnopth.png}};
			\node[inner sep=0pt,anchor=north] (cmap\pthct) at (mean\pthct.south) {\includegraphics[]{figures/RI/visualizations/\pth/colorbar_avg_recon_\imnopth.eps}};
			\node[above] at (mean\pthct.north) {\parbox{\mywidthcm}{\centering \pthtitle\\ (\pthNFEs)}};
			\coordinate (spypoint\pthct) at ($(mean\pthct.north west) + (0.9*\mywidthcm, -0.65*\mywidthcm)$);
			\coordinate  (spypointviewer\pthct) at ($(mean\pthct.north west) + (\mywidthcm, -0.2*\mywidthcm)$);
			\spy  on (spypoint\pthct) in node [fill=white] at (spypointviewer\pthct);

\node[inner sep=1pt](\pth-sd8) at (0.5*\mywidth + \pthct *\mywidth + 0.45*\pthct,-1.5*\mywidth) {\includegraphics[width=\mywidthcm]{figures/RI/visualizations_8/\pth/std_recon_\imnopth.png}};
			\node[inner sep=0pt,anchor=north] (\pth-sdcmap8) at (\pth-sd8.south) {\includegraphics[]{figures/RI/visualizations_8/\pth/colorbar_std_recon_\imnopth.eps}};
			\node[anchor = north west, inner sep=2pt] (\pth-sdcor8) at (\pth-sd8.north west) {\textcolor{white}{\input{figures/RI/visualizations_8/\pth/corr_err_std_\imnopth.txt}}};
			\coordinate (spypointsd8\pthct) at ($(\pth-sd8.north west) + (0.9*\mywidthcm, -0.65*\mywidthcm)$);
			\coordinate  (spypointviewersd8\pthct) at ($(\pth-sd8.north west) + (\mywidthcm, -0.2*\mywidthcm)$);
			\spy  on (spypointsd8\pthct) in node [fill=white] at (spypointviewersd8\pthct);

\node[inner sep=1pt](\pth-err8) at (0.5*\mywidth + \pthct *\mywidth + 0.45*\pthct,-3*\mywidth) {\includegraphics[width=\mywidthcm]{figures/RI/visualizations_8/\pth/err_\imnopth.png}};
			\node[inner sep=0pt,anchor=north] (\pth-errcmap8) at (\pth-err8.south) {\includegraphics[]{figures/RI/visualizations_8/\pth/colorbar_err_\imnopth.eps}};
			\coordinate (spypointerr8\pthct) at ($(\pth-err8.north west) + (0.9*\mywidthcm, -0.65*\mywidthcm)$);
			\coordinate  (spypointviewererr8\pthct) at ($(\pth-err8.north west) + (\mywidthcm, -0.2*\mywidthcm)$);
			\spy  on (spypointerr8\pthct) in node [fill=white] at (spypointviewererr8\pthct);
			
			\ifthenelse{\pthct<2}{
				\node[left] at (mean\pthct.west) {\rotatebox[]{90}{Mean}};
				\node[left] at (\pth-sd8.west) {\rotatebox[]{90}{\parbox{\mywidthcm}{\centering Residual \\ Pred. \\ $32\times32$}}};
				\node[left] at (\pth-err8.west) {\rotatebox[]{90}{\parbox{\mywidthcm}{\centering Residual \\ True\\ $32\times32$}}};
			}{}
			\end{scope}
		} 
	\end{tikzpicture}
\caption{Qualitative comparison of example reconstructions on ground-truth images from the PROBES dataset. The left column displays the ground truth, the pseudo-inverse $A^\dagger y$, and the sampled mask $\rvm$ applied in the Fourier domain.  We compare the posterior mean along, the residual and predicted residual (approximated via the standard dev. of 8 posterior samples) on a $32\times 32$ scale. In the upper-left corner of each predicted residual, we report the Pearson correlation to the true residual.}
	\label{fig:numerics/RI_probes_main}
\end{figure}

A qualitative comparison for a typical test image in the PROBES dataset can be found in \Cref{fig:numerics/RI_probes_main}. For each reconstruction, we plot an approximation of the posterior mean along with standard deviation and residual error plots. Due to the presence of noise in the observation $y$, pixel-wise comparisons between the standard deviation and error are challenging. To draw more meaningful comparisons, we downsample the resolution for the standard deviation and residuals by a factor of $8\times 8$. On each of the downsampled standard deviation plots, we display the Pearson correlation to the downsampled  residuals in the upper-left corner.  The zero-shot LATINO scheme finds a good approximation of large-scale features in the ground truth data, but is unable to reconstruct finer details. In contrast, the unfolded scheme, along with each of the other methods, is able to capture more fine-grained details. The IRIS schemes concentrate samples close to the posterior mean, with standard deviation and errors on a smaller scale compared to other methods. The posterior samples of U-LATINO cover attain the closest correlation to the residual error on both scales, indicating the method is able to capture areas of uncertainty in the reconstruction. Additional qualitative results, including higher-resolution comparisons of residuals, can be found in \Cref{app:ri_qualitative}.

Metric results for the PROBES data can be found in \Cref{tab:RI_4h_results}.  Compared to the zero-shot LATINO method, the unfolded model improves significantly in all metrics. U-LATINO reports the lowest LPIPS and CFID scores among all comparisons, which is expected since we include an LPIPS component in the training objective for the unfolded model using the same VGG-16 embedding as CFID.  Compared to the zero-shot IRIS SBM, we obtain comparable PSNR to IRIS after 64 E-M steps; but with 8 NFEs per sample in comparison to 64. With 1,000 E-M steps, IRIS displays a significantly PSNR around 3 decibels higher, but with a reduction in perceptual metrics and CFID. This indicates that IRIS samples may be concentrated around the posterior mean, with over-smoothing resulting in good PSNR but with lower perceptual quality and uncertainty quantification capabilities. We observe this property qualitatively in \Cref{fig:numerics/RI_probes_main}, where the standard deviation and error of IRIS samples vary only on small scales, indicating limited diversity between samples. 
When compared to the few-shot sampler, RIGAN displays improved structural qualities in the SSIM metric for both datasets. Additionally, RIGAN is the cheapest comparison, requiring a single NFE per sample. However, U-LATINO improves significantly in PSNR, LPIPS and CFID indicating improved prediction capability and posterior accuracy at a cost of requiring 8 times the number of NFEs per posterior sample. 

\Cref{tab:RI_4h_results} further displays the computational requirements to sample each model.
Due to the small number of NFEs required, (U-)LATINO and RIGAN generate a single posterior sample in less than 1 second. Since the U-LATINO was trained to generate several (correlated) samples from layers $L_0\le \ell\le L$, the time per sample was divided by a factor of 6 to reflect the time taken to compute a single sample used in the PSNR calculation. Despite the improved PSNR performance of IRIS (1000), each sample takes a factor of around 700 times longer to compute when compared to U-LATINO. LATINO and IRIS use the same foundational SBM and have a comparable number of pre-trained model weights. U-LATINO contains $\approx 3\times 10^5$ extra weights arising from unfolded hyper-parameters and LoRA fine-tuning within attention layers. LATINO and IRIS have comparable RAM usage since they rely upon the same foundational SBM prior, with LATINO having marginally lower RAM usage due to the proximal gradient steps being split into two sub-iterations.  Requiring a single NFE, RIGAN has the fastest compute time but contains around a factor 2 more model weights, leading to larger memory usage per sample. 

In summary, U-LATINO displays improved posterior quality compared to zero-shot comparisons, with significantly improved sample efficiency compared to a zero-shot SBM, at a cost of requiring an offline training phase. The unfolded MCMC architecture can be viewed as a physically interpretable neural-network, providing sufficiently improved posterior quality over a black-box conditional GAN in the PSNR, LPIPS and CFID metrics at the expense of an increased sampling cost. In \Cref{app:ri_ood_op}, we test the robustness of U-LATINO to samples of the mask $\rvm$ which fall out-of-distribution compared to the training data. 
\begin{table} 
	\centering

\begin{tabular}{|c |  c c c c c| c c c|}
			\hline
			Model (NFEs) & \makecell{PSNR \\ (Sample)} & \makecell{PSNR \\ (Mean)} & LPIPS & SSIM & CFID & \makecell{Time\\ (s)} & \makecell{RAM \\ (MB)} & \makecell{Weights\\ (M)}\\
			\hline
			RIGAN (1) & 41.51 & 43.77 & 0.07 & \bf 0.91 & 0.52 & \bf 0.04 & 1078 & 195.8\\
			\rowcolor{gray!12} U-LATINO (8) & 43.40 & 45.61 & \bf 0.01  & 0.88 & \bf 0.06 & 0.08 & \bf 683 & 88.3\\
			\hline 
			LATINO (8) & 37.33 & 37.90 & 0.08 & 0.64 & 0.19 & 0.48 & \bf 683 & 88.0\\
			IRIS (1000) & \bf 48.08 & \bf 48.99 & 0.07 & 0.86 & 2.24 & 55.93 & 718 & 88.0 \\
			IRIS (64) & 41.15 & 46.09 & 0.07 & 0.82 & 4.06 & 3.91 & 718& 88.0\\
			\hline
		\end{tabular}
	
	\caption{Reconstruction metrics for the radio-interferometry problem averaged over samples from the PROBES dataset which were not used to fine-tune our model, but which were used in pre-training the SBM in \cite{dia2025iris}. For both datasets, we compare PSNR for a single sample and for the estimated posterior mean (using the ergodic mean of $L-L_0=6$ samples for U-LATINO and 8 independent posterior samples for all other methods), LPIPS, SSIM and CFID (using a single posterior sample for each reconstruction). The average time per sample, GPU RAM and number of model parameter weights used to compute a single posterior sample are shown in the final column.} 
	\label{tab:RI_4h_results}
\end{table} 

\section{Conclusion}
In this paper, we introduced a novel framework for constructing and training deep posterior samplers by unfolding $L$  iterations of a desired Markov kernel. This provides a modular and interpretable framework for constructing conditional generative networks. At the expense of an offline training phase over a constrained class of inverse problems, unfolded MCMC architectures can fine-tune weights in a parametrised prior distribution and the Markov transition kernel to attain superior sampling accuracy and computational efficiency over traditional (zero-shot) MCMC methods. The methodology in \Cref{sec:methodology} applies generally to a broad class of MCMC transition kernels and inverse problems. We illustrate the efficiency of unfolded algorithms in two settings: an unfolded split Gibbs sampler for deblurring handwritten digits and an unfolded Langevin sampler with a score-matching prior applied to radio interferometry. Future directions of this research involve adapting this methodology to the pruning and distillation of conditional score-based models to support accurate posterior sampling with the efficiency of generative adversarial networks. A key assumption in this work is that models are trained with access to ground-truth data. An important direction of future research is to adapt the methodology to train models in a self-supervised manner, bypassing the need for ground-truth \cite{levac2025normalization}.

\paragraph{Acknowledgements}
This work was supported by UKRI Engineering and Physical Sciences Research
Council (EPSRC) (EP/V006134/1, EP/Z534481/1). We acknowledge the use of the Heriot-Watt University high-performance
computing facility (DMOG) and associated support services in the completion of this work.

\appendix

\section{Implementation Details}
\label{app:metricscompare}
\subsection{Sliced Wasserstein}
\label{app:metricscomparesw}
The sliced Wasserstein metric between $p_{\rvx,\rvy}$ and $p_{\rvx_\theta^\rvy, \rvy}$ can be expressed as 
\begin{linenomath*}\begin{equation*}
	\text{SW}(p_{\rvx,\rvy}, p_{\rvx_\theta^\rvy, \rvy}) = \E_{\pi\sim \mathcal S(\dx\times\dy)}[\mathcal W_2(\pi_\# p_{\rvx,\rvy}, \pi_\# p_{\rvx_\theta^\rvy, \rvy})],
\end{equation*}\end{linenomath*}
where $\mathcal S(\dx\times\dy)$ denotes a uniform measure on the unit sphere embedded in $\xspace\times\yspace$. We make two approximations to estimate the SW distance: firstly, for the outer expectation by using Monte Carlo approximation over 1,000 independent samples from $S(\dx\times\dy)$. Secondly, for each model posterior $\pi_\theta(x|y)$, we replace $p_{\rvx,\rvy}$ and $p_{\rvx_\theta^\rvy, \rvy}$ with empirical approximations each using $5,000$ disjoint sets of independent samples from the 10,000 test images. Approximating the $W_2$ distance between the empirical distributions instead induces a bias. We estimate the degree of bias induced by computing the SW between two empirical measures formed of independent samples sets from $p_{\rvx, \rvy}$ of size 5000. Over 10 repetitions, the baseline SW was found to be $(5.57\pm0.51)\times 10^{-5}$. As SW scores approach this value, random perturbations make similar models indistinguishable using this metric.

\subsection{Frechet Inception Distance with VAE Embedding}
\label{app:fid}
The Fr\'echet Inception Distance (FID) \cite{heusel2017gans} between $p_\rvx$ and $p_{\rvx_\theta^\rvy}$ computes the Fr\'echet distance between the two measures under an inception embedding $\mathcal E$:
\begin{linenomath*}\begin{equation*}
\text{FID}\big(p_{\rvx}, p_{\rvx_\theta^\rvy}\big) = \mathcal W_2(\mathcal E_\# p_\rvx, \mathcal E_\# p_{\rvx_\theta^\rvy}).
\end{equation*}\end{linenomath*} 
The CFID \cite{bendel2023regularized} instead computes the expected FID of posterior measures over all realisations of $\rvy$: 
\begin{linenomath*}\begin{equation*}
\text{CFID}(p_{\rvx, \rvy}, p_{\rvx_\theta^\rvy, \rvy}) = \E_{\rvy\sim p_\rvy}[\mathcal W_2(\mathcal E_\# p_{\rvx | \rvy}, \mathcal E_\# p_{\rvx_\theta^\rvy | \rvy})].
\end{equation*}\end{linenomath*} 
Typically, the FID is computed using either the Inception-V3 or VGG encoder $\mathcal E$ and uses an implicit assumption that the push-forward $\mathcal E_\# p_\rvx$ is approximately Gaussian. The mean and covariance of the embeddings are approximated by samples from the test dataset. The Fr\'echet distance between the two Gaussian approximations can then be computed exactly. Due to limitations in the Inception-V3 and VGG encoders at embedding MNIST data into a Gaussian distribution, we follow \cite[Appendix D.1]{mbakam2025learning} and  use $\mathcal E = E_\theta^{\text{VAE}}$ with the pre-trained VAE-encoder from the VAE-SGS method discussed in \Cref{sec:numerics/mnist}. Since this encoder is trained explicitly to minimise the Kullbach-Leibler divergence between $(E_\theta^{\text{VAE}})_\# p_\rvx$ and a Gaussian measure, the resulting (C)FID scores are informative. We refer to the CFID metric with $\mathcal E = E_\theta^{\text{VAE}}$ as $\mathcal W_2^\text{Latent}$ in \Cref{sec:numerics/mnist}.

\section{Additional Experimental Results}
We present here additional numerical results to highlight further properties of the unfolded models designed in \Cref{sec:numerics}. A particular emphasis is placed on exploring the adaptability of unfolded architectures to small perturbations in the observation model; a key motivation for using physics-inspired unfolded architectures.

\subsection{Out-of-Distribution Performance on MNIST}
\label{app:mnist_ood}
\begin{table}
	\centering
		\begin{tabular}{|c | c c c | c  c c|}
			\hline
			Method (NFEs) & \makecell{PSNR \\(sample)} & \makecell{PSNR \\ (mean)}  & LPIPS & \makecell{SW \\ ($\times 10^5$)}  & \makecell{$\mathcal W_2^\text{Latent}$ \\ ($\times 10^2$)} & \makecell{CMMD \\ ($\times 10^3$)} \\
			\hline
			\rowcolor{gray!12} U-SGS (4) & 22.11 & 23.55  & 0.013 & 10.30  & 4.95 & 23.44 \\
			\rowcolor{gray!12}  U-SGS (8) & 23.14 & 24.76  & 0.010 & 9.14  & 3.95 & 11.72 \\
			\rowcolor{gray!12}  U-SGS (16) & 23.78 & 25.59  & 0.009 & 7.67  & 2.16 & 17.58 \\
			\rowcolor{gray!12}  U-SGS (32) & 25.19 & 26.73  & 0.006 & \textbf{7.01}  & 0.81 & 7.81 \\
			\rowcolor{gray!12}  U-SGS (64) & \textbf{25.85} & \textbf{27.02}  & \textbf{0.005} & 7.17  & \textbf{0.54} & \textbf{6.83} \\
			\hline
			RCGAN (1) & 21.94 & 24.81  & 0.015 & 7.71  & 11.46 & 57.63 \\
			\hline
			VAE-SGS ($10^4$) & 22.76 & 22.81  & 0.017 & 13.28  & 2.26 & 18.57 \\
			\hline
		\end{tabular}
	
	\caption{Reconstruction, perceptual, and inception metrics on a single posterior sample for the Random Motion deblurring problem with out-of-distribution observation kernel $\rvk\sim \mathcal{GP}(19, \lambda_\text{Matern} = 0.5, \sigma_\text{Matern}=0.4)$.}
	\label{tab:mnist_metrics_ood_blur}
\end{table}

\begin{figure}
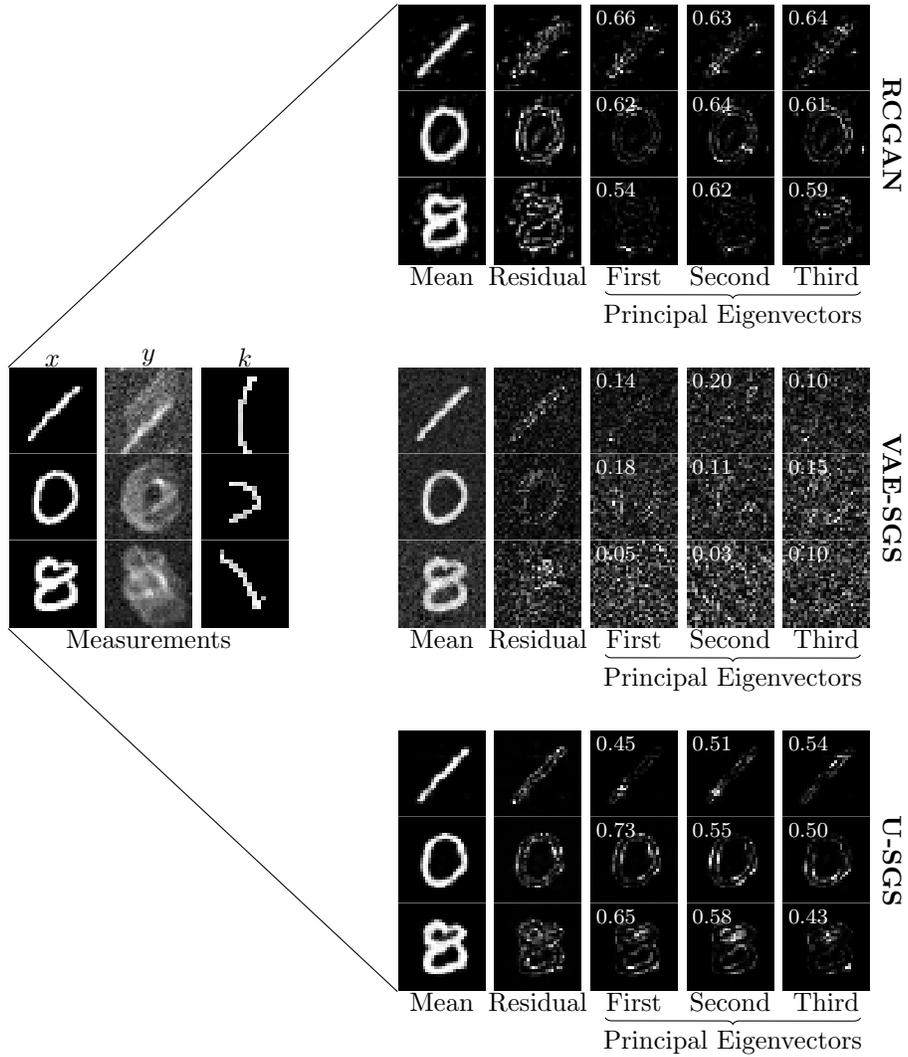

	\centering
	\begin{tikzpicture}[every node/.append style={rectangle, draw=white, thick, inner sep=0pt,}]
		
		\foreach \imno [count=\imcounter] in {11,13,15}{
			\node[inner sep = 0pt] (x\imcounter) at (-\mywidthmnist,-\imcounter * \mywidthmnist -3.65* \mywidthmnist) {\includegraphics[width=\mywidthmnistcm, height=\mywidthmnistcm]{figures/MNIST_Deblurring/visualizations/USGS_ood/ground_truth_\imno.png}};
			\node[inner sep = 0pt, right=0.1cm] (y\imcounter) at (x\imcounter.east) {\includegraphics[width=\mywidthmnistcm, height=\mywidthmnistcm]{figures/MNIST_Deblurring/visualizations/USGS_ood/measurements_\imno.png}};
			\node[inner sep = 0pt, right=0.1cm] (f\imcounter) at (y\imcounter.east) {\includegraphics[width=\mywidthmnistcm, height=\mywidthmnistcm] {figures/MNIST_Deblurring/visualizations/USGS_ood/filter_\imno.png}};
			
			\foreach \methodno [count=\methodcounter] in {RCGAN_MNIST_ood,VAE_MNIST_ood,USGS_ood}{
				
				\node[inner sep = 0pt] (m\imcounter\methodcounter) at (3.5 * \mywidthmnist,-4.2*\methodcounter* \mywidthmnist -\imcounter * \mywidthmnist + \mywidthmnist  + 3.75*\mywidthmnist) {\includegraphics[width=\mywidthmnistcm, height=\mywidthmnistcm]{figures/MNIST_Deblurring/visualizations/\methodno/avg_recon_\imno.png}};
				
				\node[inner sep = 0pt, anchor=west] (e\imcounter\methodcounter) at ($(m\imcounter\methodcounter.east) + (0.1,0)$) {\includegraphics[width=\mywidthmnistcm, height=\mywidthmnistcm]{figures/MNIST_Deblurring/visualizations/\methodno/err_\imno.png}};
				
				\node[inner sep = 0pt, anchor=west, right=0.1cm] (s\imcounter\methodcounter) at (e\imcounter\methodcounter.east) {\includegraphics[width=\mywidthmnistcm, height=\mywidthmnistcm]{figures/MNIST_Deblurring/visualizations/\methodno/principle_eigencomponent_0_\imno.png}};
				
				\node[inner sep = 2pt, anchor=north west,draw=none,] (corr\imcounter\methodcounter) at (s\imcounter\methodcounter.north west) {\textcolor{white}{\footnotesize \input{figures/MNIST_Deblurring/visualizations/\methodno/principle_eigencomponent_0_\imno.txt}}};
				
				\node[inner sep = 0pt, anchor=west, right=0.1cm] (s2\imcounter\methodcounter) at (s\imcounter\methodcounter.east) {\includegraphics[width=\mywidthmnistcm, height=\mywidthmnistcm]{figures/MNIST_Deblurring/visualizations/\methodno/principle_eigencomponent_1_\imno.png}};
				
				\node[inner sep = 2pt, anchor=north west,draw=none] (corr2\imcounter\methodcounter) at (s2\imcounter\methodcounter.north west) {\textcolor{white}{\footnotesize \input{figures/MNIST_Deblurring/visualizations/\methodno/principle_eigencomponent_1_\imno.txt}}};

				\node[inner sep = 0pt, anchor=west, right=0.1cm] (s3\imcounter\methodcounter) at (s2\imcounter\methodcounter.east) {\includegraphics[width=\mywidthmnistcm, height=\mywidthmnistcm]{figures/MNIST_Deblurring/visualizations/\methodno/principle_eigencomponent_2_\imno.png}};
				
				\node[inner sep = 2pt, anchor=north west,draw=none] (corr3\imcounter\methodcounter) at (s3\imcounter\methodcounter.north west) {\textcolor{white}{\footnotesize \input{figures/MNIST_Deblurring/visualizations/\methodno/principle_eigencomponent_2_\imno.txt}}};
				
				\ifthenelse{\imcounter>2}{
					\node[anchor=north, below] (m\methodcounter_title) at (m3\methodcounter.south) {Mean}; 
					\node[anchor=north, below] (e\methodcounter_title) at (e3\methodcounter.south) {Residual}; 
					\node[anchor=north, below] (s\methodcounter_title) at (s3\methodcounter.south) {First}; 
					\node[anchor=north, below] (s2\methodcounter_title) at (s23\methodcounter.south) {Second};
					\node[anchor=north, below] (s3\methodcounter_title) at (s33\methodcounter.south) {Third};
					\draw[decoration={brace,mirror,raise=0.5ex}, anchor=north,decorate] (s\methodcounter_title.south west) -- node [below=1.5ex,anchor=north] {Principal Eigenvectors} (s3\methodcounter_title.south east);
				}{}
				
			}
			
		}

		\node[anchor=south, above] (mlab1) at (x1.north) {$x$};
		\node[anchor=south,above] (mlab2) at (y1.north) {$y$};
		\node[anchor=south,above] (mlab3) at (f1.north) {$k$};
		
		\node[anchor=west,right=1ex] (slab1) at (s321.east) {\rotatebox{270}{\bf RCGAN}};
		\node[anchor=west,right = 1ex] (slab2) at (s323.east) {\rotatebox{270}{\bf U-SGS}};
		\node[anchor=west, right=1ex] (slab3) at (s322.east) {\rotatebox{270}{\bf VAE-SGS}};
		
\node[anchor=north, below] (ytitle) at (y3.south) {Measurements}; 
		
		\draw [] (x1.north west) -- (m11.north west);
		\draw [] (x3.south west) -- (m33.south west);
		
	\end{tikzpicture}
	\caption{Qualitative comparison for motion deblurring applied to the MNIST dataset with out-of-distribution kernels $\rvk\sim \mathcal{GP}(19, \lambda_\text{Matern} = 0.5, \sigma_\text{Matern}=0.4)$.}
	\label{fig:qualitative_comparison_MNIST_ood}
\end{figure}

 The U-SGS model trained in \Cref{sec:numerics/mnist} contains a natural embedding through proximal gradient steps of the forward operator $\op$. Consequently, the trained model was able to  well approximate the task of motion deblurring where $\op$ was sampled randomly from a class of motion blur kernels. In this section, we present numerical results which evaluate the performance of the pre-trained models on out-of-training-distribution inverse problems by randomly sampling the motion blur kernel using a different generating distribution $\rvk\sim \mathcal{GP}(19, \lambda_\text{Matern} = 0.5, \sigma_\text{Matern}=0.4)$ during the test loop. Kernels from this distribution cover a larger receptive field with greater variability. We present quantitative results in \Cref{tab:mnist_metrics_ood_blur}. All methods report worse performance than in \Cref{tab:mnist_metrics}, reflecting the more challenging forward operation. However, the difference in performance between each U-SGS model and RCGAN is greater than the experiment on in-distribution inverse problems. This can be observed qualitatively in \Cref{fig:qualitative_comparison_MNIST_ood} where the artefacts arising in samples from the black-box RCGAN due to the random forward operators are amplified for the out-of-distribution problem. In contrast, U-SGS and the zero-shot VAE-SGS methods both generate more realistic samples from the MNIST dataset.

\subsection{Additional Results on PROBES data}
\label{app:ri_qualitative}
In this supplement, we provide further qualitative and quantitative results for the RI reconstructions discussed in \Cref{sec:numerics/RI}.

\subsubsection{Residuals at Higher Resolution}
\begin{figure}
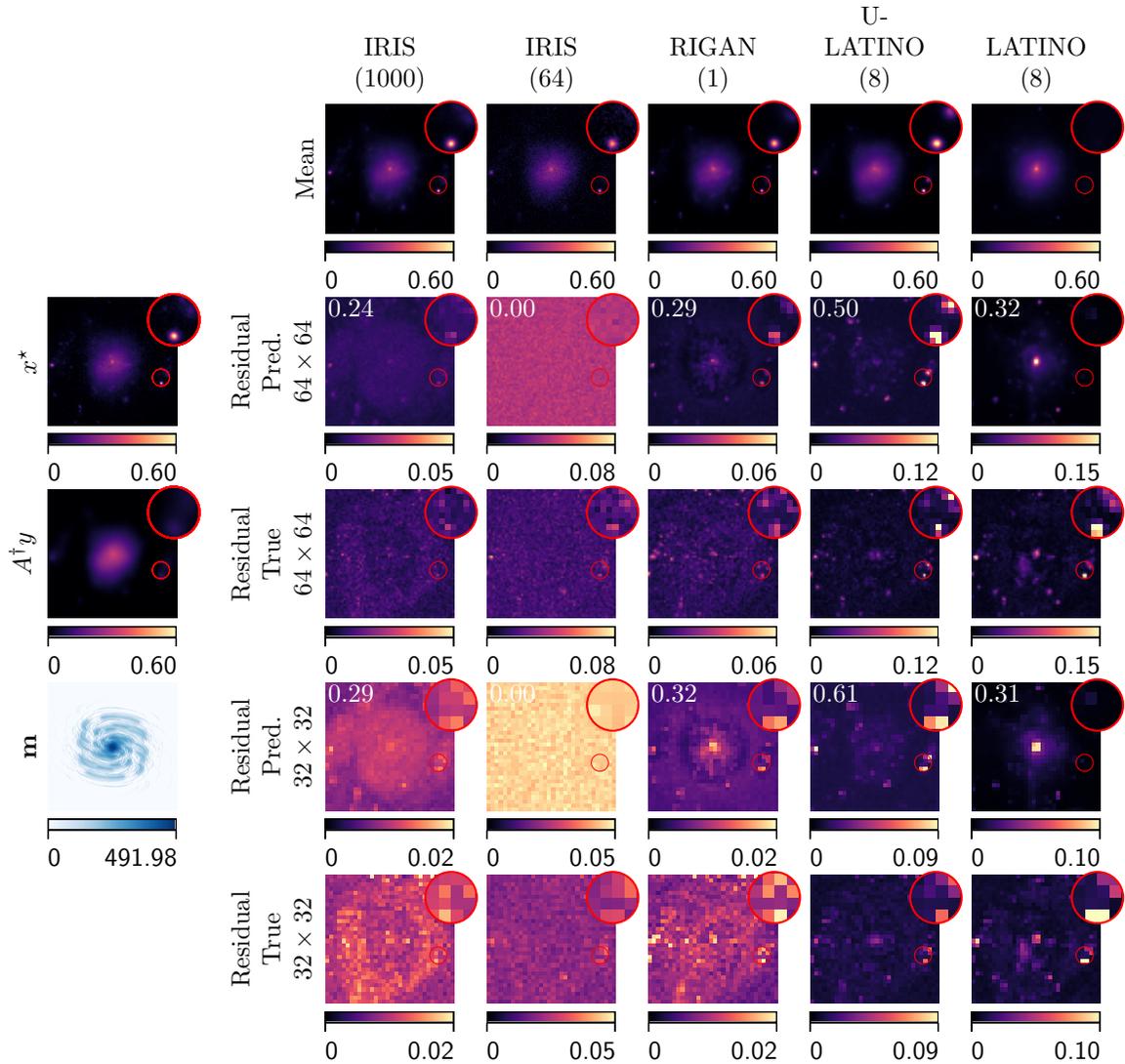

	\centering
	\begin{tikzpicture}[spy using outlines={circle, red, size=0.4*\mywidthcm, magnification=3}]
		\node[inner sep = 1pt](x1) at (-0.4*\mywidth,-1.5*\mywidth) {\includegraphics[width=\mywidthcm]{figures/RI/\visdir/ULAT_probes_4h/ground_truth_4.png}};
		\node[inner sep=0pt,anchor=north] (x1cmap) at (x1.south) {\includegraphics[]{figures/RI/\visdir/ULAT_probes_4h/colorbar_ground_truth_4.eps}};
		\node[left] at (x1.west) {\rotatebox[]{90}{$x^\star$}};
		
		\node[inner sep=1pt] (y1) at (-0.4*\mywidth,-3*\mywidth) {\includegraphics[width=\mywidthcm]{figures/RI/\visdir/ULAT_probes_4h/measurements_4.png}};
		\node[inner sep=0pt,anchor=north] (y1cmap) at (y1.south) {\includegraphics[]{figures/RI/\visdir/ULAT_probes_4h/colorbar_measurements_4.eps}};
		\node[left] at (y1.west) {\rotatebox[]{90}{$A^\dagger y$}};
		
		\node[inner sep = 1pt](m1) at (-0.4*\mywidth,-4.5*\mywidth) {\includegraphics[width=\mywidthcm]{figures/RI/\visdir/ULAT_probes_4h/mask_4.png}};
		\node[inner sep=0pt,anchor=north] (m1cmap) at (m1.south) {\includegraphics[]{figures/RI/\visdir/ULAT_probes_4h/colorbar_mask_4.eps}};
		\node[left] at (m1.west) {\rotatebox[]{90}{$\rvm$}};
		\coordinate (spypoint) at ($(x1.north west) + (0.9*\mywidthcm, -0.65*\mywidthcm)$);
		\coordinate  (spypointviewer) at ($(x1.north west) + (\mywidthcm, -0.2*\mywidthcm)$);
		\spy  on (spypoint) in node [fill=white] at (spypointviewer);
		
		\coordinate (spypointy) at ($(y1.north west) + (0.9*\mywidthcm, -0.65*\mywidthcm)$);
		\coordinate  (spypointyviewer) at ($(y1.north west) + (\mywidthcm, -0.2*\mywidthcm)$);
		\spy  on (spypointy) in node [fill=white] at (spypointyviewer);

		\foreach \pth/\pthtitle/\pthNFEs/\imnopth [count=\pthct from 1] in {IRIS_probes_4h_1000_steps_img4/IRIS/1000/0, IRIS_probes_4h_64_steps_img4/IRIS/64/0, RCGAN_probes_4h/RIGAN/1/4, ULAT_probes_4h/U-LATINO/8/4, LAT_probes_4h/LATINO/8/4}{
			\begin{scope}[spy using outlines={circle, red, size=0.4*\mywidthcm, magnification=3}]
				\node[inner sep=1pt](mean\pthct) at (0.5*\mywidth + \pthct *\mywidth + 0.45*\pthct,0) {\includegraphics[width=\mywidthcm]{figures/RI/visualizations/\pth/avg_recon_\imnopth.png}};
				\node[inner sep=0pt,anchor=north] (cmap\pthct) at (mean\pthct.south) {\includegraphics[]{figures/RI/visualizations/\pth/colorbar_avg_recon_\imnopth.eps}};
				\node[above] at (mean\pthct.north) {\parbox{\mywidthcm}{\centering \pthtitle\\ (\pthNFEs)}};
				\coordinate (spypoint\pthct) at ($(mean\pthct.north west) + (0.9*\mywidthcm, -0.65*\mywidthcm)$);
				\coordinate  (spypointviewer\pthct) at ($(mean\pthct.north west) + (\mywidthcm, -0.2*\mywidthcm)$);
				\spy  on (spypoint\pthct) in node [fill=white] at (spypointviewer\pthct);
				
				\node[inner sep=1pt](\pth-sd) at (0.5*\mywidth + \pthct *\mywidth + 0.45*\pthct,-1.5*\mywidth) {\includegraphics[width=\mywidthcm]{figures/RI/visualizations/\pth/std_recon_\imnopth.png}};
				\node[inner sep=0pt,anchor=north] (\pth-sdcmap) at (\pth-sd.south) {\includegraphics[]{figures/RI/visualizations/\pth/colorbar_std_recon_\imnopth.eps}};
				\node[anchor = north west, inner sep=2pt] (\pth-sdcor) at (\pth-sd.north west) {\textcolor{white}{\input{figures/RI/visualizations/\pth/corr_err_std_\imnopth.txt}}};
				\coordinate (spypointsd4\pthct) at ($(\pth-sd.north west) + (0.9*\mywidthcm, -0.65*\mywidthcm)$);
				\coordinate  (spypointviewersd4\pthct) at ($(\pth-sd.north west) + (\mywidthcm, -0.2*\mywidthcm)$);
				\spy  on (spypointsd4\pthct) in node [fill=white] at (spypointviewersd4\pthct);
				
				\node[inner sep=1pt](\pth-err) at (0.5*\mywidth + \pthct *\mywidth + 0.45*\pthct,-3*\mywidth) {\includegraphics[width=\mywidthcm]{figures/RI/visualizations/\pth/err_\imnopth.png}};
				\node[inner sep=0pt,anchor=north] (\pth-errcmap) at (\pth-err.south) {\includegraphics[]{figures/RI/visualizations/\pth/colorbar_err_\imnopth.eps}};
				\coordinate (spypointerr4\pthct) at ($(\pth-err.north west) + (0.9*\mywidthcm, -0.65*\mywidthcm)$);
				\coordinate  (spypointviewererr4\pthct) at ($(\pth-err.north west) + (\mywidthcm, -0.2*\mywidthcm)$);
				\spy  on (spypointerr4\pthct) in node [fill=white] at (spypointviewererr4\pthct);

				\node[inner sep=1pt](\pth-sd8) at (0.5*\mywidth + \pthct *\mywidth + 0.45*\pthct,-4.5*\mywidth) {\includegraphics[width=\mywidthcm]{figures/RI/visualizations_8/\pth/std_recon_\imnopth.png}};
				\node[inner sep=0pt,anchor=north] (\pth-sdcmap8) at (\pth-sd8.south) {\includegraphics[]{figures/RI/visualizations_8/\pth/colorbar_std_recon_\imnopth.eps}};
				\node[anchor = north west, inner sep=2pt] (\pth-sdcor8) at (\pth-sd8.north west) {\textcolor{white}{\input{figures/RI/visualizations_8/\pth/corr_err_std_\imnopth.txt}}};
				\coordinate (spypointsd8\pthct) at ($(\pth-sd8.north west) + (0.9*\mywidthcm, -0.65*\mywidthcm)$);
				\coordinate  (spypointviewersd8\pthct) at ($(\pth-sd8.north west) + (\mywidthcm, -0.2*\mywidthcm)$);
				\spy  on (spypointsd8\pthct) in node [fill=white] at (spypointviewersd8\pthct);

				\node[inner sep=1pt](\pth-err8) at (0.5*\mywidth + \pthct *\mywidth + 0.45*\pthct,-6*\mywidth) {\includegraphics[width=\mywidthcm]{figures/RI/visualizations_8/\pth/err_\imnopth.png}};
				\node[inner sep=0pt,anchor=north] (\pth-errcmap8) at (\pth-err8.south) {\includegraphics[]{figures/RI/visualizations_8/\pth/colorbar_err_\imnopth.eps}};
				\coordinate (spypointerr8\pthct) at ($(\pth-err8.north west) + (0.9*\mywidthcm, -0.65*\mywidthcm)$);
				\coordinate  (spypointviewererr8\pthct) at ($(\pth-err8.north west) + (\mywidthcm, -0.2*\mywidthcm)$);
				\spy  on (spypointerr8\pthct) in node [fill=white] at (spypointviewererr8\pthct);
				
				\ifthenelse{\pthct<2}{
					\node[left] at (mean\pthct.west) {\rotatebox[]{90}{Mean}};
					\node[left] at (\pth-sd.west) {\rotatebox[]{90}{\parbox{\mywidthcm}{\centering Residual \\ Pred. \\ $64\times64$}}};
					\node[left] at (\pth-err.west) {\rotatebox[]{90}{\parbox{\mywidthcm}{\centering Residual \\ True\\ $64\times64$}}};
					\node[left] at (\pth-sd8.west) {\rotatebox[]{90}{\parbox{\mywidthcm}{\centering Residual \\ Pred. \\ $32\times32$}}};
					\node[left] at (\pth-err8.west) {\rotatebox[]{90}{\parbox{\mywidthcm}{\centering Residual \\ True\\ $32\times32$}}};
				}{}
			\end{scope}
		} 
	\end{tikzpicture}
	\caption{Qualitative comparison of an example reconstruction on ground-truth images from the PROBES dataset. The left column displays the ground truth data, the pseudo-inverse of $A$ applied to $y$, and the sampled mask $\rvm$ applied in the Fourier domain.  For each method, we compare the posterior mean, standard deviation and the absolute error computed between the posterior mean and ground truth.}
	\label{fig:numerics/RI_probes_4}
\end{figure}

To compare the ability of each model to quantify uncertainties at different scales, in \Cref{fig:numerics/RI_probes_4} we display the reconstructions in \Cref{fig:numerics/RI_probes_main} with additional residuals at a scale of $64\times64$ pixels. Qualitative results on a different image from the test set can be found in \Cref{fig:numerics/RI_probes_4}. At the higher resolution, LATINO and IRIS (64) are again unable to capture and quantify uncertainty around finer structures in the ground-truth image, with U-LATINO displaying errors focused on directions aligned with the ground truth image and samples which have a standard deviation highly correlated to the residuals. 

\begin{figure}
	\centering
	\begin{tikzpicture}[spy using outlines={circle, red, size=0.4*\mywidthcm, magnification=3}]
		\node[inner sep = 1pt](x1) at (-0.4*\mywidth,-1.5*\mywidth) {\includegraphics[width=\mywidthcm]{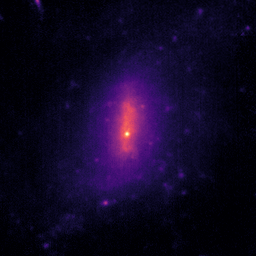}};
		\node[inner sep=0pt,anchor=north] (x1cmap) at (x1.south) {\includegraphics[]{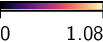}};
		\node[left] at (x1.west) {\rotatebox[]{90}{$x^\star$}};
		
		\node[inner sep=1pt] (y1) at (-0.4*\mywidth,-3*\mywidth) {\includegraphics[width=\mywidthcm]{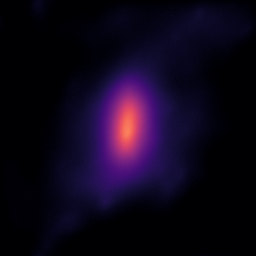}};
		\node[inner sep=0pt,anchor=north] (y1cmap) at (y1.south) {\includegraphics[]{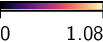}};
		\node[left] at (y1.west) {\rotatebox[]{90}{$A^\dagger y$}};
		
		\node[inner sep = 1pt](m1) at (-0.4*\mywidth,-4.5*\mywidth) {\includegraphics[width=\mywidthcm]{figures/RI/\visdir/ULAT_probes_4h/mask_4.png}};
		\node[inner sep=0pt,anchor=north] (m1cmap) at (m1.south) {\includegraphics[]{figures/RI/\visdir/ULAT_probes_4h/colorbar_mask_4.eps}};
		\node[left] at (m1.west) {\rotatebox[]{90}{$\rvm$}};
		\coordinate (spypoint) at ($(x1.north west) + (0.45*\mywidthcm, -0.8*\mywidthcm)$);
		\coordinate  (spypointviewer) at ($(x1.north west) + (\mywidthcm, -0.2*\mywidthcm)$);
		\spy  on (spypoint) in node [fill=white] at (spypointviewer);
		
		\coordinate (spypointy) at ($(y1.north west) + (0.45*\mywidthcm, -0.8*\mywidthcm)$);
		\coordinate  (spypointyviewer) at ($(y1.north west) + (\mywidthcm, -0.2*\mywidthcm)$);
		\spy  on (spypointy) in node [fill=white] at (spypointyviewer);

		\foreach \pth/\pthtitle/\pthNFEs/\imnopth [count=\pthct from 1] in {IRIS_probes_4h_1000_steps/IRIS/1000/0, IRIS_probes_4h_64_steps/IRIS/64/0, RCGAN_probes_4h/RIGAN/1/13, ULAT_probes_4h/U-LATINO/8/13, LAT_probes_4h/LATINO/8/13}{
			\begin{scope}[spy using outlines={circle, red, size=0.4*\mywidthcm,  magnification=3}]
				\node[inner sep=1pt](mean\pthct) at (0.5*\mywidth + \pthct *\mywidth + 0.45*\pthct,0) {\includegraphics[width=\mywidthcm]{figures/RI/visualizations/\pth/avg_recon_\imnopth.png}};
				\node[inner sep=0pt,anchor=north] (cmap\pthct) at (mean\pthct.south) {\includegraphics[]{figures/RI/visualizations/\pth/colorbar_avg_recon_\imnopth.eps}};
				\node[above] at (mean\pthct.north) {\parbox{\mywidthcm}{\centering \pthtitle\\ (\pthNFEs)}};
				\coordinate (spypoint\pthct) at ($(mean\pthct.north west) + (0.45*\mywidthcm, -0.8*\mywidthcm)$);
				\coordinate  (spypointviewer\pthct) at ($(mean\pthct.north west) + (\mywidthcm, -0.2*\mywidthcm)$);
				\spy  on (spypoint\pthct) in node [fill=white] at (spypointviewer\pthct);
				
				\node[inner sep=1pt](\pth-sd) at (0.5*\mywidth + \pthct *\mywidth + 0.45*\pthct,-1.5*\mywidth) {\includegraphics[width=\mywidthcm]{figures/RI/visualizations/\pth/std_recon_\imnopth.png}};
				\node[inner sep=0pt,anchor=north] (\pth-sdcmap) at (\pth-sd.south) {\includegraphics[]{figures/RI/visualizations/\pth/colorbar_std_recon_\imnopth.eps}};
				\node[anchor = north west, inner sep=2pt] (\pth-sdcor) at (\pth-sd.north west) {\textcolor{white}{\input{figures/RI/visualizations/\pth/corr_err_std_\imnopth.txt}}};
				\coordinate (spypointsd4\pthct) at ($(\pth-sd.north west) + (0.45*\mywidthcm, -0.8*\mywidthcm)$);
				\coordinate  (spypointviewersd4\pthct) at ($(\pth-sd.north west) + (\mywidthcm, -0.2*\mywidthcm)$);
				\spy  on (spypointsd4\pthct) in node [fill=white] at (spypointviewersd4\pthct);
				
				\node[inner sep=1pt](\pth-err) at (0.5*\mywidth + \pthct *\mywidth + 0.45*\pthct,-3*\mywidth) {\includegraphics[width=\mywidthcm]{figures/RI/visualizations/\pth/err_\imnopth.png}};
				\node[inner sep=0pt,anchor=north] (\pth-errcmap) at (\pth-err.south) {\includegraphics[]{figures/RI/visualizations/\pth/colorbar_err_\imnopth.eps}};
				\coordinate (spypointerr4\pthct) at ($(\pth-err.north west) + (0.45*\mywidthcm, -0.8*\mywidthcm)$);
				\coordinate  (spypointviewererr4\pthct) at ($(\pth-err.north west) + (\mywidthcm, -0.2*\mywidthcm)$);
				\spy  on (spypointerr4\pthct) in node [fill=white] at (spypointviewererr4\pthct);

				\node[inner sep=1pt](\pth-sd8) at (0.5*\mywidth + \pthct *\mywidth + 0.45*\pthct,-4.5*\mywidth) {\includegraphics[width=\mywidthcm]{figures/RI/visualizations_8/\pth/std_recon_\imnopth.png}};
				\node[inner sep=0pt,anchor=north] (\pth-sdcmap8) at (\pth-sd8.south) {\includegraphics[]{figures/RI/visualizations_8/\pth/colorbar_std_recon_\imnopth.eps}};
				\node[anchor = north west, inner sep=2pt] (\pth-sdcor8) at (\pth-sd8.north west) {\textcolor{white}{\input{figures/RI/visualizations_8/\pth/corr_err_std_\imnopth.txt}}};
				\coordinate (spypointsd8\pthct) at ($(\pth-sd8.north west) + (0.45*\mywidthcm, -0.8*\mywidthcm)$);
				\coordinate  (spypointviewersd8\pthct) at ($(\pth-sd8.north west) + (\mywidthcm, -0.2*\mywidthcm)$);
				\spy  on (spypointsd8\pthct) in node [fill=white] at (spypointviewersd8\pthct);

				\node[inner sep=1pt](\pth-err8) at (0.5*\mywidth + \pthct *\mywidth + 0.45*\pthct,-6*\mywidth) {\includegraphics[width=\mywidthcm]{figures/RI/visualizations_8/\pth/err_\imnopth.png}};
				\node[inner sep=0pt,anchor=north] (\pth-errcmap8) at (\pth-err8.south) {\includegraphics[]{figures/RI/visualizations_8/\pth/colorbar_err_\imnopth.eps}};
				\coordinate (spypointerr8\pthct) at ($(\pth-err8.north west) + (0.45*\mywidthcm, -0.8*\mywidthcm)$);
				\coordinate  (spypointviewererr8\pthct) at ($(\pth-err8.north west) + (\mywidthcm, -0.2*\mywidthcm)$);
				\spy  on (spypointerr8\pthct) in node [fill=white] at (spypointviewererr8\pthct);
				
				\ifthenelse{\pthct<2}{
					\node[left] at (mean\pthct.west) {\rotatebox[]{90}{Mean}};
					\node[left] at (\pth-sd.west) {\rotatebox[]{90}{\parbox{\mywidthcm}{\centering Residual \\ Pred. \\ $64\times64$}}};
					\node[left] at (\pth-err.west) {\rotatebox[]{90}{\parbox{\mywidthcm}{\centering Residual \\ True\\ $64\times64$}}};
					\node[left] at (\pth-sd8.west) {\rotatebox[]{90}{\parbox{\mywidthcm}{\centering Residual \\ Pred. \\ $32\times32$}}};
					\node[left] at (\pth-err8.west) {\rotatebox[]{90}{\parbox{\mywidthcm}{\centering Residual \\ True\\ $32\times32$}}};
				}{}
			\end{scope}
		} 
	\end{tikzpicture}
	\caption{Qualitative comparison of an example reconstruction on ground-truth images from the PROBES dataset. The left column displays the ground truth data, the pseudo-inverse of $A$ applied to $y$, and the sampled mask $\rvm$ applied in the Fourier domain.  For each method, we compare the posterior mean, standard deviation and the absolute error computed between the posterior mean and ground truth.}
	\label{fig:numerics/RI_probes}
\end{figure}

\subsubsection{MCMC Sample Diversity}	
\begin{figure}
	\centering 
	\begin{tikzpicture}
		\begin{scope}[spy using outlines={circle, red, magnification=3, size=0.4*\mywidthcm}]
			\node[inner sep = 1pt](x1) at (1.5*\mywidth + 1.5*0.25*\mywidth,0) {\includegraphics[width=\mywidthcm]{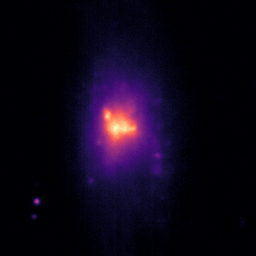}};
			\node[inner sep=0pt,anchor=north] (x1cmap) at (x1.south) {\includegraphics[]{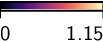}};
			\node[above] at (x1.north) {\rotatebox[]{0}{$x^\star$}};
			\coordinate (spypt) at ($(x1.north west) + (0.15*\mywidthcm, -0.8*\mywidthcm)$);
			\coordinate (spyv) at ($(x1.north west) + (\mywidthcm, -0.2*\mywidthcm)$);
			\spy  on (spypt) in node [fill=white] at (spyv);
			
			\node[inner sep = 1pt](y1) at (2.5*\mywidth + 2.5*0.25*\mywidth,0) {\includegraphics[width=\mywidthcm]{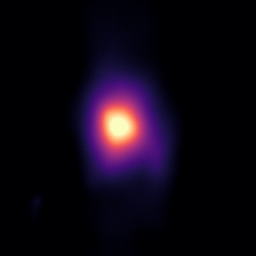}};
			\node[inner sep=0pt,anchor=north] (y1cmap) at (y1.south) {\includegraphics[]{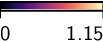}};
			\node[above] at (y1.north) {\rotatebox[]{0}{$A^\dagger x$}};
			\coordinate (spypty) at ($(y1.north west) + (0.15*\mywidthcm, -0.8*\mywidthcm)$);
			\coordinate (spyvy) at ($(y1.north west) + (\mywidthcm, -0.2*\mywidthcm)$);
			\spy  on (spypty) in node [fill=white] at (spyvy);
			
			\node[inner sep = 1pt](m1) at (3.5*\mywidth + 3.5*0.25*\mywidth,0) {\includegraphics[width=\mywidthcm]{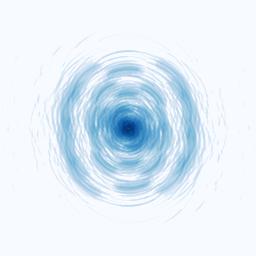}};
			\node[inner sep=0pt,anchor=north] (m1cmap) at (m1.south) {\includegraphics[]{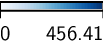}};
			\node[above] at (m1.north) {\rotatebox[]{0}{$\rvm$}};
		\end{scope}
		\def\ktwo{3}
		\foreach \k in {0,...,5}{
			\begin{scope}[spy using outlines={circle, red, magnification=3, size=0.4*\mywidthcm}]
				\node[inner sep = 1pt](x\k1) at (\k*\mywidth +0.25*\k*\mywidth,-2*\mywidth) {\includegraphics[width=\mywidthcm]{figures/RI/visualizations/ULAT_probes_4h/mcmc_sample_\k_12.png}};
				\node[inner sep=0pt,anchor=north] (x\k1cmap) at (x\k1.south) {\includegraphics[]{figures/RI/visualizations/ULAT_probes_4h/colorbar_mcmc_sample_\k_12.eps}};
				\node[above] at (x\k1.north) {\rotatebox[]{0}{$\ell=\pgfmathparse{int(\k+\ktwo)}\pgfmathresult$}};
				\coordinate (spypt\k) at ($(x\k1.north west) + (0.15*\mywidthcm, -0.8*\mywidthcm)$);
				\coordinate (spyv\k) at ($(x\k1.north west) + (\mywidthcm, -0.2*\mywidthcm)$);
				\spy  on (spypt\k) in node [fill=white] at (spyv\k);
				
				\node[inner sep = 1pt](x2\k1) at (\k*\mywidth +0.25*\k*\mywidth,-3.5*\mywidth) {\includegraphics[width=\mywidthcm]{figures/RI/visualizations/LAT_probes_4h/mcmc_sample_\k_12.png}};
				\node[inner sep=0pt,anchor=north] (x2\k1cmap) at (x2\k1.south) {\includegraphics[]{figures/RI/visualizations/LAT_probes_4h/colorbar_mcmc_sample_\k_12.eps}};
				\coordinate (spypt2\k) at ($(x2\k1.north west) + (0.15*\mywidthcm, -0.8*\mywidthcm)$);
				\coordinate (spyv2\k) at ($(x2\k1.north west) + (\mywidthcm, -0.2*\mywidthcm)$);
				\spy  on (spypt2\k) in node [fill=white] at (spyv2\k);
			\end{scope}
		}
		\node[left] at (x01.west) {\rotatebox[]{90}{U-LATINO}};
		\node[left] at (x201.west) {\rotatebox[]{90}{LATINO}};
		
	\end{tikzpicture}
	\caption{Output of final 6 MCMC samples comparing U-LATINO to LATINO. The ground truth $x^\star$ is taken from the PROBES hold-out validation data and observation kernel $\rvk\in K_\text{4h}$.} 
	\label{fig:app/ri_mcmc_samples}
\end{figure}

Both LATINO and the corresponding unfolded model rely on the same underlying Markovian sampling kernel.  In \Cref{fig:app/ri_mcmc_samples}, we investigate the diversity of the (correlated) samples from a single run of the MCMC chain, following the burn-in phase. We see that the zero-shot LATINO exhibits limited diversity between samples, with a low-quality final approximation of the ground-truth image. In contrast, the unfolded model exhibits fast convergence to the true data, and shows significant sample diversity around areas of uncertainty in the measurement data. A zoomed in region shows a fine-detail area of the ground-truth, which is captured in the unfolded model but not explored by the zero-shot method. 
\subsubsection{Comparison on SBM Prior Data}
\begin{figure}
	\centering
	\begin{tikzpicture}[spy using outlines={circle, red, size=0.4*\mywidthcm,  magnification=3}]
		\node[inner sep = 1pt](x1) at (-0.4*\mywidth,-1.5*\mywidth) {\includegraphics[width=\mywidthcm]{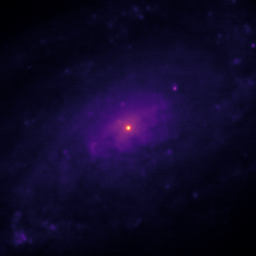}};
		\node[inner sep=0pt,anchor=north] (x1cmap) at (x1.south) {\includegraphics[]{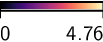}};
		\node[left] at (x1.west) {\rotatebox[]{90}{$x^\star$}};
		
		\node[inner sep=1pt] (y1) at (-0.4*\mywidth,-3*\mywidth) {\includegraphics[width=\mywidthcm]{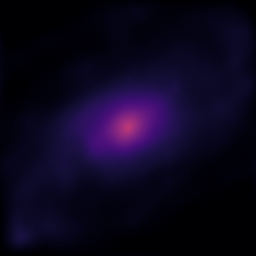}};
		\node[inner sep=0pt,anchor=north] (y1cmap) at (y1.south) {\includegraphics[]{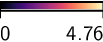}};
		\node[left] at (y1.west) {\rotatebox[]{90}{$A^\dagger y$}};
		
		\node[inner sep = 1pt](m1) at (-0.4*\mywidth,-4.5*\mywidth) {\includegraphics[width=\mywidthcm]{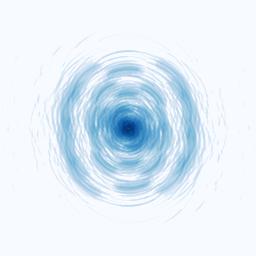}};
		\node[inner sep=0pt,anchor=north] (m1cmap) at (m1.south) {\includegraphics[]{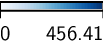}};
		\node[left] at (m1.west) {\rotatebox[]{90}{$\rvm$}};
		
		\coordinate (spypoint) at ($(x1.north west) + (0.7*\mywidthcm, -0.4*\mywidthcm)$);
		\coordinate  (spypointviewer) at ($(x1.north west) + (\mywidthcm, -0.2*\mywidthcm)$);
		\spy  on (spypoint) in node [fill=white] at (spypointviewer);
		
		\coordinate (spypointy) at ($(y1.north west) + (0.7*\mywidthcm, -0.4*\mywidthcm)$);
		\coordinate  (spypointyviewer) at ($(y1.north west) + (\mywidthcm, -0.2*\mywidthcm)$);
		\spy  on (spypointy) in node [fill=white] at (spypointyviewer);

		\foreach \pth/\pthtitle/\pthNFEs/\imnopth [count=\pthct from 1] in {IRIS_sbm_4h_1000_steps/IRIS/1000/0, IRIS_sbm_4h_64_steps/IRIS/64/0, RCGAN_diff_4h/RIGAN/1/12, ULAT_diff_probes_4h/U-LATINO/8/12, LAT_diff_probes_4h/LATINO/8/12}{
			\begin{scope}[spy using outlines={circle, red, size=0.4*\mywidthcm,  magnification=3}]
				\node[inner sep=1pt](mean\pthct) at (0.5*\mywidth + \pthct *\mywidth + 0.45*\pthct,0) {\includegraphics[width=\mywidthcm]{figures/RI/visualizations/\pth/avg_recon_\imnopth.png}};
				\node[inner sep=0pt,anchor=north] (cmap\pthct) at (mean\pthct.south) {\includegraphics[]{figures/RI/visualizations/\pth/colorbar_avg_recon_\imnopth.eps}};
				\node[above] at (mean\pthct.north) {\parbox{\mywidthcm}{\centering \pthtitle\\ (\pthNFEs)}};
				\coordinate (spypoint\pthct) at ($(mean\pthct.north west) + (0.7*\mywidthcm, -0.4*\mywidthcm)$);
				\coordinate  (spypointviewer\pthct) at ($(mean\pthct.north west) + (\mywidthcm, -0.2*\mywidthcm)$);
				\spy  on (spypoint\pthct) in node [fill=white] at (spypointviewer\pthct);
				
				\node[inner sep=1pt](\pth-sd) at (0.5*\mywidth + \pthct *\mywidth + 0.45*\pthct,-1.5*\mywidth) {\includegraphics[width=\mywidthcm]{figures/RI/visualizations/\pth/std_recon_\imnopth.png}};
				\node[inner sep=0pt,anchor=north] (\pth-sdcmap) at (\pth-sd.south) {\includegraphics[]{figures/RI/visualizations/\pth/colorbar_std_recon_\imnopth.eps}};
				\node[anchor = north west, inner sep=2pt] (\pth-sdcor) at (\pth-sd.north west) {\textcolor{white}{\input{figures/RI/visualizations/\pth/corr_err_std_\imnopth.txt}}};
				\coordinate (spypointsd4\pthct) at ($(\pth-sd.north west) + (0.7*\mywidthcm, -0.4*\mywidthcm)$);
				\coordinate  (spypointviewersd4\pthct) at ($(\pth-sd.north west) + (\mywidthcm, -0.2*\mywidthcm)$);
				\spy  on (spypointsd4\pthct) in node [fill=white] at (spypointviewersd4\pthct);
				
				\node[inner sep=1pt](\pth-err) at (0.5*\mywidth + \pthct *\mywidth + 0.45*\pthct,-3*\mywidth) {\includegraphics[width=\mywidthcm]{figures/RI/visualizations/\pth/err_\imnopth.png}};
				\node[inner sep=0pt,anchor=north] (\pth-errcmap) at (\pth-err.south) {\includegraphics[]{figures/RI/visualizations/\pth/colorbar_err_\imnopth.eps}};
				\coordinate (spypointerr4\pthct) at ($(\pth-err.north west) + (0.7*\mywidthcm, -0.4*\mywidthcm)$);
				\coordinate  (spypointviewererr4\pthct) at ($(\pth-err.north west) + (\mywidthcm, -0.2*\mywidthcm)$);
				\spy  on (spypointerr4\pthct) in node [fill=white] at (spypointviewererr4\pthct);

				\node[inner sep=1pt](\pth-sd8) at (0.5*\mywidth + \pthct *\mywidth + 0.45*\pthct,-4.5*\mywidth) {\includegraphics[width=\mywidthcm]{figures/RI/visualizations_8/\pth/std_recon_\imnopth.png}};
				\node[inner sep=0pt,anchor=north] (\pth-sdcmap8) at (\pth-sd8.south) {\includegraphics[]{figures/RI/visualizations_8/\pth/colorbar_std_recon_\imnopth.eps}};
				\node[anchor = north west, inner sep=2pt] (\pth-sdcor8) at (\pth-sd8.north west) {\textcolor{white}{\input{figures/RI/visualizations_8/\pth/corr_err_std_\imnopth.txt}}};
				\coordinate (spypointsd8\pthct) at ($(\pth-sd8.north west) + (0.7*\mywidthcm, -0.4*\mywidthcm)$);
				\coordinate  (spypointviewersd8\pthct) at ($(\pth-sd8.north west) + (\mywidthcm, -0.2*\mywidthcm)$);
				\spy  on (spypointsd8\pthct) in node [fill=white] at (spypointviewersd8\pthct);

				\node[inner sep=1pt](\pth-err8) at (0.5*\mywidth + \pthct *\mywidth + 0.45*\pthct,-6*\mywidth) {\includegraphics[width=\mywidthcm]{figures/RI/visualizations_8/\pth/err_\imnopth.png}};
				\node[inner sep=0pt,anchor=north] (\pth-errcmap8) at (\pth-err8.south) {\includegraphics[]{figures/RI/visualizations_8/\pth/colorbar_err_\imnopth.eps}};
				\coordinate (spypointerr8\pthct) at ($(\pth-err8.north west) + (0.7*\mywidthcm, -0.4*\mywidthcm)$);
				\coordinate  (spypointviewererr8\pthct) at ($(\pth-err8.north west) + (\mywidthcm, -0.2*\mywidthcm)$);
				\spy  on (spypointerr8\pthct) in node [fill=white] at (spypointviewererr8\pthct);
				
				\ifthenelse{\pthct<2}{
					\node[left] at (mean\pthct.west) {\rotatebox[]{90}{Mean}};
					\node[left] at (\pth-sd.west) {\rotatebox[]{90}{\parbox{\mywidthcm}{\centering Residual \\ Pred. \\ $64\times64$}}};
					\node[left] at (\pth-err.west) {\rotatebox[]{90}{\parbox{\mywidthcm}{\centering Residual \\ True\\ $64\times64$}}};
					\node[left] at (\pth-sd8.west) {\rotatebox[]{90}{\parbox{\mywidthcm}{\centering Residual \\ Pred. \\ $32\times32$}}};
					\node[left] at (\pth-err8.west) {\rotatebox[]{90}{\parbox{\mywidthcm}{\centering Residual \\ True\\ $32\times32$}}};
				}{}
			\end{scope}
		} 
	\end{tikzpicture}
	\caption{Qualitative comparison of an example reconstruction on synthetic ground-truth validation images sampled from the score-based model prior. The left column displays the ground truth data, the pseudo-inverse of $A$ applied to $y$, and the sampled mask $\rvm$ applied in the Fourier domain.  For each method, we compare the posterior mean, standard deviation and the absolute error computed between the posterior mean and ground truth. To ease visual comparison, the mask $\rvm$, the standard deviation and the absolute error are visualised on a log-scale colour gradient.}
	\label{fig:numerics/RI_sbm}
\end{figure}

As a further benchmark for the models considered in \Cref{sec:numerics/RI}, we present here a separate comparison using a synthetic PROBES test set. For this data, we generate 128 ground-truth images from the IRIS SBM prior for the purpose of testing each model. A qualitative comparison for a typical reconstruction from this data is shown in \Cref{fig:numerics/RI_sbm}, and quantitative metrics, averaged over the test set are shown in \Cref{tab:app/sbmquantitative}. For the quantitative results, we observe a slight improvement in all metrics for each method compared to \Cref{tab:RI_4h_results}. This can be attributed to less background noise in the SBM prior data, along with the IRIS and LATINO models using a prior $p_\theta$ which matches exactly the generating distribution for images in this test set. Compared to U-LATINO, the zero-shot LATINO attains a higher SSIM score and correlation between posterior standard-deviation and residual error on this dataset image. We attribute this to the zero-shot model over-fitting to the prior data, combined with the fact that U-LATINO was not fine-tuned on images from this SBM prior.

\begin{table}
	\centering
	\begin{tabular}{|c | c c c c|}
		\hline
		Model (NFEs) & PSNR  & LPIPS & SSIM & CFID\\
		\hline
		RIGAN (1) & 43.22 & 0.06 & 0.91 & 0.48\\
		\rowcolor{gray!12} U-LATINO (8) & 44.97 & \bf 0.01  & 0.79 & \bf 0.02\\
		\hline
		LATINO (8) & 39.00 & 0.03 & \bf 0.92 & 0.08\\
		IRIS (1000) & \bf 50.87 & 0.05 & 0.87 & 0.08\\
		IRIS (64) & 44.36 & 0.08 & 0.75 & 2.45\\
		\hline
	\end{tabular}
	\caption{Reconstruction metrics for the RI task using the PROBES score-based prior images.}
	\label{tab:app/sbmquantitative}
\end{table}

\subsection{Out-of-Distribution Masks for Radio Interferometry}
\label{app:ri_ood_op}
In \Cref{app:mnist_ood}, we observed that unfolded networks extend naturally to small perturbations in the observation process. We test a similar setup for the RI task here. In particular, we consider the same forward observation mode \eqref{eqn:numerics/RI_task} as in \Cref{sec:numerics/RI} but for each observation generating a mask $\rvm \sim \mathcal U(\mathcal f K_\text{2h})$ using 2-hour visibility patterns instead of 4-hour observations. 
The out-of-distribution visibilities contain less coverage in high-frequency components, reflecting a slightly more challenging reconstruction task. We test each model on the PROBES test set using the new observation masks. Neither U-LATINO nor RIGAN was fine-tuned to fit masks from $\mathcal f K_\text{2h}$. The remaining methods are zero-shot and designed to operate for any linear Gaussian observation model.

\begin{figure}
	\centering
	\begin{tikzpicture}[spy using outlines={circle, red, size=0.4*\mywidthcm,  magnification=3}]
\node[inner sep = 1pt](x1) at (-0.4*\mywidth,-1.5*\mywidth) {\includegraphics[width=\mywidthcm]{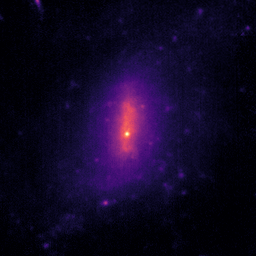}};
		\node[inner sep=0pt,anchor=north] (x1cmap) at (x1.south) {\includegraphics[]{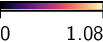}};
		\node[left] at (x1.west) {\rotatebox[]{90}{$x^\star$}};
		
		\node[inner sep=1pt] (y1) at (-0.4*\mywidth,-3*\mywidth) {\includegraphics[width=\mywidthcm]{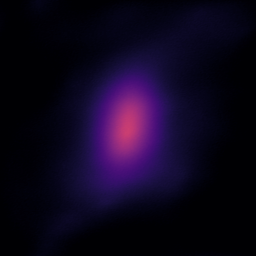}};
		\node[inner sep=0pt,anchor=north] (y1cmap) at (y1.south) {\includegraphics[]{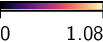}};
		\node[left] at (y1.west) {\rotatebox[]{90}{$A^\dagger y$}};
		
		\node[inner sep = 1pt](m1) at (-0.4*\mywidth,-4.5*\mywidth) {\includegraphics[width=\mywidthcm]{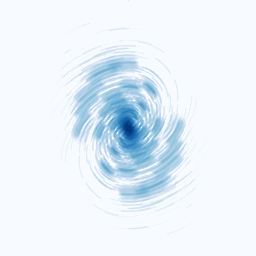}};
		\node[inner sep=0pt,anchor=north] (m1cmap) at (m1.south) {\includegraphics[]{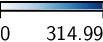}};
		\node[left] at (m1.west) {\rotatebox[]{90}{$\rvm$}};
\coordinate (spypoint) at ($(x1.north west) + (0.45*\mywidthcm, -0.8*\mywidthcm)$);
		\coordinate  (spypointviewer) at ($(x1.north west) + (\mywidthcm, -0.2*\mywidthcm)$);
		\spy  on (spypoint) in node [fill=white] at (spypointviewer);
		
		\coordinate (spypointy) at ($(y1.north west) + (0.45*\mywidthcm, -0.8*\mywidthcm)$);
		\coordinate  (spypointyviewer) at ($(y1.north west) + (\mywidthcm, -0.2*\mywidthcm)$);
		\spy  on (spypointy) in node [fill=white] at (spypointyviewer);
		
		\foreach \pth/\pthtitle/\pthNFEs/\imnopth [count=\pthct from 1] in {IRIS_probes_2h_1000_steps/IRIS/1000/0, IRIS_probes_2h_64_steps/IRIS/64/0, RCGAN_probes_2h/RIGAN/1/13, ULAT_probes_2h/U-LATINO/8/13, LAT_probes_2h/LATINO/8/13}{
			\begin{scope}[spy using outlines={circle, red, size=0.4*\mywidthcm,  magnification=3}]
\node[inner sep=1pt](mean\pthct) at (0.5*\mywidth + \pthct *\mywidth + 0.45*\pthct,0) {\includegraphics[width=\mywidthcm]{figures/RI/visualizations/\pth/avg_recon_\imnopth.png}};
				\node[inner sep=0pt,anchor=north] (cmap\pthct) at (mean\pthct.south) {\includegraphics[]{figures/RI/visualizations/\pth/colorbar_avg_recon_\imnopth.eps}};
				\node[above] at (mean\pthct.north) {\parbox{\mywidthcm}{\centering \pthtitle\\ (\pthNFEs)}};
				\coordinate (spypoint\pthct) at ($(mean\pthct.north west) + (0.45*\mywidthcm, -0.8*\mywidthcm)$);
				\coordinate  (spypointviewer\pthct) at ($(mean\pthct.north west) + (\mywidthcm, -0.2*\mywidthcm)$);
				\spy  on (spypoint\pthct) in node [fill=white] at (spypointviewer\pthct);
				
\node[inner sep=1pt](\pth-sd) at (0.5*\mywidth + \pthct *\mywidth + 0.45*\pthct,-1.5*\mywidth) {\includegraphics[width=\mywidthcm]{figures/RI/visualizations/\pth/std_recon_\imnopth.png}};
				\node[inner sep=0pt,anchor=north] (\pth-sdcmap) at (\pth-sd.south) {\includegraphics[]{figures/RI/visualizations/\pth/colorbar_std_recon_\imnopth.eps}};
				\node[anchor = north west, inner sep=2pt] (\pth-sdcor) at (\pth-sd.north west) {\textcolor{white}{\input{figures/RI/visualizations/\pth/corr_err_std_\imnopth.txt}}};
				\coordinate (spypointsd4\pthct) at ($(\pth-sd.north west) + (0.45*\mywidthcm, -0.8*\mywidthcm)$);
				\coordinate  (spypointviewersd4\pthct) at ($(\pth-sd.north west) + (\mywidthcm, -0.2*\mywidthcm)$);
				\spy  on (spypointsd4\pthct) in node [fill=white] at (spypointviewersd4\pthct);
				
\node[inner sep=1pt](\pth-err) at (0.5*\mywidth + \pthct *\mywidth + 0.45*\pthct,-3*\mywidth) {\includegraphics[width=\mywidthcm]{figures/RI/visualizations/\pth/err_\imnopth.png}};
				\node[inner sep=0pt,anchor=north] (\pth-errcmap) at (\pth-err.south) {\includegraphics[]{figures/RI/visualizations/\pth/colorbar_err_\imnopth.eps}};
				\coordinate (spypointerr4\pthct) at ($(\pth-err.north west) + (0.45*\mywidthcm, -0.8*\mywidthcm)$);
				\coordinate  (spypointviewererr4\pthct) at ($(\pth-err.north west) + (\mywidthcm, -0.2*\mywidthcm)$);
				\spy  on (spypointerr4\pthct) in node [fill=white] at (spypointviewererr4\pthct);

\node[inner sep=1pt](\pth-sd8) at (0.5*\mywidth + \pthct *\mywidth + 0.45*\pthct,-4.5*\mywidth) {\includegraphics[width=\mywidthcm]{figures/RI/visualizations_8/\pth/std_recon_\imnopth.png}};
				\node[inner sep=0pt,anchor=north] (\pth-sdcmap8) at (\pth-sd8.south) {\includegraphics[]{figures/RI/visualizations_8/\pth/colorbar_std_recon_\imnopth.eps}};
				\node[anchor = north west, inner sep=2pt] (\pth-sdcor8) at (\pth-sd8.north west) {\textcolor{white}{\input{figures/RI/visualizations_8/\pth/corr_err_std_\imnopth.txt}}};
				\coordinate (spypointsd8\pthct) at ($(\pth-sd8.north west) + (0.45*\mywidthcm, -0.8*\mywidthcm)$);
				\coordinate  (spypointviewersd8\pthct) at ($(\pth-sd8.north west) + (\mywidthcm, -0.2*\mywidthcm)$);
				\spy  on (spypointsd8\pthct) in node [fill=white] at (spypointviewersd8\pthct);

\node[inner sep=1pt](\pth-err8) at (0.5*\mywidth + \pthct *\mywidth + 0.45*\pthct,-6*\mywidth) {\includegraphics[width=\mywidthcm]{figures/RI/visualizations_8/\pth/err_\imnopth.png}};
				\node[inner sep=0pt,anchor=north] (\pth-errcmap8) at (\pth-err8.south) {\includegraphics[]{figures/RI/visualizations_8/\pth/colorbar_err_\imnopth.eps}};
				\coordinate (spypointerr8\pthct) at ($(\pth-err8.north west) + (0.45*\mywidthcm, -0.8*\mywidthcm)$);
				\coordinate  (spypointviewererr8\pthct) at ($(\pth-err8.north west) + (\mywidthcm, -0.2*\mywidthcm)$);
				\spy  on (spypointerr8\pthct) in node [fill=white] at (spypointviewererr8\pthct);
				
				\ifthenelse{\pthct<2}{
					\node[left] at (mean\pthct.west) {\rotatebox[]{90}{Mean}};
					\node[left] at (\pth-sd.west) {\rotatebox[]{90}{\parbox{\mywidthcm}{\centering Residual \\ Pred. \\ $64\times64$}}};
					\node[left] at (\pth-err.west) {\rotatebox[]{90}{\parbox{\mywidthcm}{\centering Residual \\ True\\ $64\times64$}}};
					\node[left] at (\pth-sd8.west) {\rotatebox[]{90}{\parbox{\mywidthcm}{\centering Residual \\ Pred. \\ $32\times32$}}};
					\node[left] at (\pth-err8.west) {\rotatebox[]{90}{\parbox{\mywidthcm}{\centering Residual \\ True\\ $32\times32$}}};
				}{}
			\end{scope}
		} 
	\end{tikzpicture}
	\caption{Qualitative comparison of an example reconstruction on a ground-truth image from the PROBES dataset using an out-of-distribution kernel $\rvk\sim\mathcal{U}(K_\text{2h})$.}
	\label{fig:app/ri_ood_mask}
\end{figure}

A qualitative comparison for an example out-of distribution reconstruction can be found in \Cref{fig:app/ri_ood_mask}; quantitative results averaged over the PROBES dataset can be found in \Cref{tab:app/ri_ood_mask}. With the exception of the unfolded model U-LATINO, all methods have a slight reduction in performance compared to \Cref{tab:RI_4h_results}. In contrast, the change in forward operators does not affect the performance of U-LATINO, highlighting the resilience of unfolded architectures to subtle changes in the recovery task.

\begin{table} 
	\centering

\begin{tabular}{|c | c c c c c|}
		\hline
		Model (NFEs) & \makecell{PSNR \\ (Sample)} & \makecell{PSNR\\ (Mean)} & LPIPS & SSIM & CFID\\
		\hline
		RIGAN (1) & 40.35 & 42.19 & 0.08 & \textbf{0.91} & 0.55\\
		\rowcolor{gray!12} U-LATINO (8) & 43.57 & 45.50 & \textbf{0.02} & 0.82 & \textbf{0.06}\\
		\hline 
		LATINO (8) & 37.53 & 37.89 & 0.07 & 0.62 & 0.20\\
		IRIS (1000) & \bf 47.74 & \textbf{48.53} & 0.06 & 0.83 & 0.32\\
		IRIS (64)  & 40.94 & 45.70 & 0.08 & 0.81 & 3.22 \\
		\hline
	\end{tabular}
	\caption{Reconstruction metrics for the RI problem using out-of-distribution kernels ${\rvk\sim \mathcal U(K_\text{2h})}.$} 
	\label{tab:app/ri_ood_mask}
\end{table}

\clearpage
\bibliographystyle{abbrv}
\bibliography{references}

\end{document}